%% file: main.tex
\documentclass{article}
\usepackage[preprint]{neurips_2021}

\usepackage[utf8]{inputenc} %
\usepackage[T1]{fontenc}    %
\usepackage{hyperref}       %
\usepackage{url}            %
\usepackage{booktabs}       %
\usepackage{amsfonts}       %
\usepackage{nicefrac}       %
\usepackage{microtype}      %
\usepackage[square,sort,comma,numbers]{natbib}

\usepackage{enumerate}
\usepackage{amsmath}
\usepackage{xcolor}
\usepackage{bm}
\usepackage{graphicx}
\usepackage{float}
\usepackage{pifont}
\usepackage{hyperref}
\usepackage{graphicx}
\usepackage{subcaption}
\usepackage{mwe}
\usepackage[linesnumbered,ruled,lined,noend]{algorithm2e}

\newcommand{\cX}{\mathcal{X}}

\newcommand{\cP}{\mathcal{P}}

\renewcommand{\vec}[1]{\bm{#1}}

\SetKwProg{Fn}{function}{}{}
\SetKwInOut{Ret}{return}
\SetKwComment{Hline}{}{\vspace{-3mm}\textcolor{gray}{\hrule}\vspace{1mm}}
\SetKwRepeat{Do}{do}{while}

\title{Multi-objective Asynchronous Successive Halving}

\author{%
  Robin Schmucker\thanks{Work done while interning at Amazon, Berlin, Germany.} \\
  Machine Learning Department\\
  Carnegie Mellon University\\
  \texttt{rschmuck@cs.cmu.edu} 
    \And
    Michele Donini\\
  Amazon\\
  Berlin, Germany\\
  \texttt{donini@amazon.com} \\\\
    \And
     Muhammad Bilal Zafar\\
  Amazon\\
  Berlin, Germany\\
  \texttt{zafamuh@amazon.com} \\\\
   \And
    David Salinas\\
  Amazon\\
  Berlin, Germany\\
  \texttt{dsalina@amazon.com} \\\\
   \And
    Cédric Archambeau\\
  Amazon\\
  Berlin, Germany\\
  \texttt{cedrica@amazon.com} \\\\
}

\begin{document}
\maketitle

\input{text/abstract}

\input{text/introduction}
\input{text/related_work}

\section{Background: Multi-Objective Optimization (MO)}
\label{sec:multi_opt}

\paragraph{The MO problem}
In many real world settings there is not just one, but multiple potentially conflicting objectives of interest. For example when working with hardware constraint systems one desires predictive models that are accurate and have low resource consumption. Assume one needs to decide between two solutions $\vec{x}_1, \vec{x}_2$. Each individual objective induces a distinct ordinal relationship between the candidates, but while $\vec{x}_1$ might dominate $\vec{x}_2$ in one objective it might be worse in another. Because there is rarely a single best solution, the MO problem is to identify the Pareto front -- the set of all non-dominated solutions -- to help users to make an informed trade-off. 

More formally, let $f: \cX \rightarrow \mathbb{R}^n$ be a function over domain $\cX$ that we aim to minimize. Given two points $\vec{x}_1, \vec{x}_2 \in \cX$, we write $\vec{x}_1 \succeq \vec{x}_2$ if $\vec{x}_2$ is {\it weakly-dominated} by $\vec{x}_1$, that is, iff $f(\vec{x}_1)_i \leq f(\vec{x}_2)_i, \forall i \in [n]$.
We write $\vec{x}_1 \succ \vec{x}_2$ if $\vec{x}_2$ is {\it dominated} by $\vec{x}_1$, that is, iff $\vec{x}_1 \succeq \vec{x}_2$ and  $\exists  i \in [n]$ s.t. $f(\vec{x}_1)_i < f(\vec{x}_2)_i$. The Pareto front of $f$ is defined by $\cP_f = \{\vec{x} \in \cX  | \not \exists \vec{x}' \in  \cX : \vec{x}'  \succ \vec{x}\}$, that is, the set of all non-dominated points. As $\cP_f$ is often an infinite object, MO algorithms aim to recover an \textit{approximation set} $A \subset \cX$ of non-dominated objective vectors. A popular measure of approximation quality is the \textit{dominated hypervolume}~\cite{Zitzler1998:Multiobjective}. Given an approximation set $A$ and a {\it reference point} $\vec{r}$ the hypervolume indicator $\mathcal{H}$ is given by:
\begin{align}
\mathcal{H}(A) =\text{Vol} \left(\{\vec{x} \in \mathbb{R}^n | \exists \vec{z} \in A: \vec{z} \succeq \vec{x} \wedge \vec{x} \succeq \vec{r}\}\right).
\end{align}
Hypervolume related quantities are usually computed by partitioning the space into hyper-cubes which are then summed. This operation scales exponentially with the number objective functions and can cause a bottleneck for related MO optimization approaches. 

\paragraph{Scalarization for MO} Scalarization techniques reduce the MO to an SO problem by optimizing a linear combination of the multiple competing objectives. For some $\vec{w}\in\mathbb{R}^n$, solving the MO minimization problem consists in solving the following optimization problem:
\begin{equation}
    \min_{\vec{x} \in \cX} V(f(\vec{x}, r_t), \vec{w}) ,
\end{equation}
where $V: \mathbf{R}^n \times \mathbf{R}^n \rightarrow \mathbf{R}$ is a scalarization function of choice. Some concrete choices of $V$ are the inner product between objective and weight vector (also known an Random Weights) and the scalarization functions employed by ParEGO~\cite{Knowles2006:Parego} and Golovin~\cite{Golovin2020}:
\begin{align}
\label{eq:scalarizations}
\begin{split}
V_{\text{RW}}(\vec{y}, \vec{w}) = \vec{y}^\top \vec{w}, \quad V_{\text{ParEGO}}(\vec{y}, \vec{w}) = \max_{j \in \{1, \dots, n\}}(w_j y_j) + \rho (\vec{y}^\top \vec{w})\\
V_{\text{Golovin}}(\vec{y}, \vec{w}) = \min_{j \in \{1, \dots, n\}} (\max (0, y_j / w_j))^n. \quad \quad \quad \quad \quad
\end{split}
\end{align}

\input{text/method}

\input{text/background}
\input{text/experiments}
\input{text/conclusion}

\medskip
\bibliographystyle{plain}
\bibliography{bibliography_robin,bibliography,references}

\newpage
\appendix

\input{text/appendix}

\end{document}

%% file: text/abstract.tex
\begin{abstract}
    Hyperparameter optimization (HPO) is increasingly used to automatically tune the predictive performance (e.g., accuracy) of machine learning models. However, in a plethora of real-world applications, accuracy is only one of the multiple -- often conflicting -- performance criteria, necessitating the %
    adoption of a multi-objective (MO) perspective. While the literature on MO optimization is rich, few prior studies have focused on HPO. In this paper, we propose algorithms that extend asynchronous successive halving (ASHA) to the MO setting. 
    Considering multiple evaluation metrics, we assess the performance of these methods on three real world tasks: (i) Neural architecture search, (ii) algorithmic fairness and (iii) language model optimization. Our empirical analysis shows that MO ASHA enables to perform MO HPO at scale. Further, we observe that that taking the entire Pareto front into account for candidate selection consistently outperforms multi-fidelity HPO based on MO scalarization in terms of wall-clock time. Our algorithms (to be open-sourced) establish new baselines for future research in the area.

\end{abstract}

%% file: text/introduction.tex
\section{Introduction}
\label{sec:introduction}

With the increasing complexity of deep neural networks (DNNs), such as convolutional neural networks (CNNs)~\cite{Lecun1999:Object} and Transformers~\cite{Vaswani2017:Attention}, the task of %
tuning %
their hyperparameters %
has become critically important, but also %
time-consuming and costly. To automate this %
process, hyperparameter optimization %
(HPO) %
is %
routinely used to 
maximize predictive performance. 
However, in many real-world applications, predictive performance is \textit{not} the only relevant performance objective. Instead, %
users are interested in simultaneously optimizing \textit{several} objectives such as fairness, interpretability or latency. %
For instance, prior work has shown that improving fairness can have an adverse impact on model accuracy~\cite{corbett2017algorithmic} and that optimizing fairness w.r.t. one metric can deteriorate %
others~\cite{kleinberg2016inherent}. Similarly, while increasing the size of a LSTM-based language model can lead to improvements in perplexity, it can also significantly increase the prediction cost~\cite{Strubell2019:Energy}.

Despite the pervasiveness of multi-objective (MO) problems %
in the context of 
HPO, the area of multi-objective hyperparameter optimization (MO-HPO) has received little attention from the research community so far. While a few studies exist%
, most notably, based on multi-objective Bayesian optimization (MBO)%
, these are limited to a narrow range of tasks~\cite{Lobato2016} (i.e. fast and accurate neural networks for MNIST~\cite{Lecun1998:MNIST}) %
and the empirical analysis %
is limited; evaluations are often performed on synthetic functions and/or focuses on a small %
number of objectives, usually no more than three.

In this paper, we %
extend asynchronous successive halving (ASHA)~\cite{Li2018:Massively}, which is a scheduler for distributed multi-fidelity HPO, to the MO setting. 
As is standard when optimizing several objectives simultaneously, we cast the problem as that of approximating the Pareto front. 
The Pareto front is the set of all non-dominated solutions to the MO problem; these are the solutions for which none of the objectives can be improved without degrading another one. Recovering the Pareto is convenient in practice as it enables users to navigate the \textit{optimal} trade-off between the multiple objectives. 

Our goal %
is to efficiently recover well-rounded fronts in %
real-world scenarios. %
The gradient-free MO optimization algorithms we introduce leverage an asynchronous successive halving scheduler (ASHA)~\cite{Li2018:Massively} and rely on MO optimization strategies %
based on scalarizing the objective vector (e.g., scalarization technique of~\cite{Knowles2006:Parego, Golovin2020}), as well as approaches that %
take the geometry of the whole Pareto front into account (e.g., the non-dominated sorting technique of~\cite{Deb2002:Fast}). 
The resulting MO-HPO algorithms %
benefit from early stopping and parallelization to deliver significant performance gains over existing MBO techniques.

We evaluate the performance of these algorithms using several performance metrics such as dominated hypervolume~\cite{Zitzler1998:Multiobjective} and quality of the minimax solution. We also investigate how different algorithms navigate the solution space over time and the speed at which they recover the Pareto front. We perform an empirical comparison across a diverse range of MO tasks ranging from fairness-aware learning over neural architecture search to language modeling. Results %
indicate that techniques %
that take the geometry of the current Pareto front approximation 
into account 
perform \textit{consistently} better than the scalarization-based techniques across different tasks. In fact, in some cases, the scalarization-based techniques are not significantly better than a random search over the %
hyperparameter space.

To summarize, our main contributions are: 
\begin{enumerate}
    \item We obtain multi-fidelity HPO algorithms by extending ASHA to the MO setting, considering scalarization-based selection techniques, as well as techniques informed by the geometry of the Pareto front.
    \item We evaluate %
    these multi-fidelity MO-HPO algorithms on a range of real-world tasks and evaluation metrics, showing that %
    they can be used effectively to navigate trade-offs between various objectives.
    \item We show that exploration schemes informed by the geometry of the Pareto front outperform the scalarization based techniques on HPO problems.
\end{enumerate}

%% file: text/related_work.tex
\section{Related Work}
\label{sec:related_work}

BO allows to build cheap to evaluate surrogate models that can be used to trade-off exploration and exploitation. Multi-objective Bayesian optimization (MBO) techniques have been developed and can be categorized into three classes. First, \emph{scalarization}-based MBO methods map the vector of all objectives to a single scalar value~\cite{Knowles2006:Parego, Zhang2009:Expensive, Nakayama2009:Sequential,Paria2019:Flexible, Zhang2020:Random,Golovin2020} and then use conventional single-objective (SO) BO techniques. 
While these methods are comparatively easy to implement and scale gracefully with the number of objectives, they are not guaranteed to convergence to the full Pareto front with the notable exception of~\cite{Golovin2020}. %
A second class of MBO techniques %
optimize a performance measure of Pareto front approximations, namely the \textit{dominated hypervolume}~\cite{Zitzler1998:Multiobjective}.
Both the expected hypervolume improvement (EHI)~\cite{Emmerich2011:Hypervolume} and probability of hyper-improvement (PHI)~\cite{Keane2006:Statistical} %
extend their single-objective BO counterparts, the expected improvement~\cite{Mockus1978:Application, Jones1998:Efficient} and the probability of improvement~\cite{Kushner1964:New}, respectively.
Other methods include step-wise uncertainty reduction (SUR)~\cite{Picheny2015:Multiobjective}, smsEGO~\cite{Ponweiser2008:Multiobjective, Wagner2010:Expected} and expected maximin improvement (EMMI)~\cite{Svenson2010:Multiobjective}.
Third, \emph{information-theoretic} MBO approaches
have been proposed. They aim to %
reduce uncertainty about the location of the Pareto front. Methods in this class tend to be more sample efficient and scale better with the number of objectives. PAL~\cite{Zuluaga2013:Active} iteratively reduces the size of a discrete uncertainty set.
PESMO~\cite{Hernandez2016:Predictive} adapts the predictive entropy search (PES)~\cite{Hernandez2014:Predictive} criterion. Pareto front entropy search (PFES)~\cite{Suzuki2019:Multi} is suitable when dealing with decoupled objectives.
MESMO~\cite{Belakaria2019:Max} builds on the max-value entropy-search criterion~\cite{Wang2017:Max} and enjoys an asymptotic regret bound. Building on MESMO, two recent works have proposed MF-OSEMO~\cite{Belakaria2020:Multi} and iMOCA~\cite{Belakaria2020:Information}, two multi-fidelity based information-theoretic MBO techniques which internally use %
continuous fidelity Gaussian processes. 

MO evolutionary algorithms (EA) identify promising solution candidates that are close to the Pareto front and mutate them to conduct exploration~\cite{Elsken2019}. At the heart of many EA methods lies non-dominated-sorting~\cite{srinivas1994} which ranks candidates in a two stage process. First, candidates are grouped into multiple distinct Pareto fronts by iteratively computing a front and reducing the candidate set until all points are assigned. Second, candidates inside the individual fronts are ranked using crowding distance~\cite{Deb2002:Fast} to encourage uniform exploration. We refer to~\cite{Emmerich2018} for a survey on EA approaches.

Rather than evaluating poor models for the maximum resource budget, typically epochs, %
multi-fidelity HPO iteratively allocates resources to hyperparameter configurations that perform sufficiently well at certain checkpoints (or rungs). Hyperband~\cite{Li2017Hyperband} is an extension of Successive Halving~\cite{Jamieson2016:Non} which samples random candidates, compares their performance after their intermediate budget allocation is exhausted, and only let the most promising candidates run for larger computational budgets. %
Waiting for all configurations to finish to make a stopping decision is %
wasteful as it causes idle workers. ASHA~\cite{Li2018:Massively} is an asynchronous extension of Hyperband, yielding a large improvement in wall clock time. 
Recently, \cite{Schmucker2020:Multi} applied simple scalarization (i.e., linear mixture of objectives) to Hyperband and~\cite{Salinas2021:Multi} investigated the combination of EA and Hyperband when considering tuning hyperparameters and hardware jointly.
Finally a parallel work investigates the performance of different MF approaches when optimizing  only two objectives for computer-vision tasks~\cite{guerrero2021}.
In our work, we adapted the non-dominating sorting idea from EA with an ASHA-style early-stopping mechanism, investigating the performance of the proposed methods in several different scenarios, from NAS to Algorithmic Fairness optimizing over many different objectives.

%% file: text/method.tex
\section{Distributed Asynchronous Multi-objective Hyperparameter Optimization}
\label{sec:method}

Our work builds on \cite{Schmucker2020:Multi}. The authors evaluated various MBO techniques on MO-HPO tasks and artificial functions (an excerpt of their study is provided in Appendix~\ref{app:mbo_comparison}). They reported, perhaps surprisingly, that in many cases random search could yield a sample complexity comparable to BO in the MO setting, while being easy to parallelize. Motivated by these observations, they proposed a Hyperband-style algorithm that uses random scalarization for hyperparameter candidate selection. %
Given a computational budget, single objective Hyperband~\cite{Li2017Hyperband} starts by providing a small initial resource allocation $r_0$ to each randomly sampled model configuration. If a configuration does not seem promising after its allocation is exhausted (e.g., in terms of validation error), Hyperband uses an early-stopping rule to terminate unpromising candidates and reallocates (larger) additional resource $r_t$ to the most promising ones. This process is repeated until the total budget is exhausted.  %
Hyperband extends the %
Successive Halving algorithm~\cite{Jamieson2016:Non} by introducing an additional parameter $\eta$ which is used to trade-of the number candidate hyperparameters with a per candidate budget.

To adapt Hyperband to the MO domain, \cite{Schmucker2020:Multi} use %
scalarization %
to reduce each configuration's evaluation vector to a single real valued scalar which allows for an easy candidate ranking. For this each individual configuration $\vec{x} \in \cX$ is equipped with a distinct set of $m$ vectors $W = \{\vec{w}_1, \dots, \vec{w}_m\}$ sampled uniformly from $\Delta_n$ (i.e., the set of all vectors in the positive orthant with entries summing up to one). The performance of configuration $\vec{x}$ after partial training is then measured by
\begin{equation}
    e_V(\vec{x}, W, r_t) = \min_{\vec{w} \in W} V(f(\vec{x}, r_t), \vec{w}) .
\end{equation}
While the use of scalarization techniques is a convenient response to the candidate selection problem, as it allows the application of single objective BO, it comes with multiple shortcomings. It is known that scalarization techniques can only capture restricted Pareto shapes~\cite{Golovin2020}. Moreover, each individual weight vector only encourages exploration in a single direction and, if the set of weight vectors $W$ is not chosen carefully, it can fail to recover the full Pareto front altogether. The use of linear scalarization schemes also tends to focus heavily on the objective value of smallest magnitude which can cause failure for non-convex Pareto fronts. Finally, scalarization techniques are heavily affected by rescaling of objective values.
\begin{figure*}[t]
    \centering
\begin{minipage}{0.52\textwidth}
\begin{algorithm}[H]\small
\caption{Multi-objective selector}
\label{algo:nondominated}
\DontPrintSemicolon
\SetAlgoLined
\BlankLine
\Fn{\normalfont\texttt{non\_dom\_sorting}($P$)}{
    $i = 1$\;
    \While{\normalfont $P \neq \emptyset$}{
        $F_i \gets \{ \vec{\theta} \in P: \vec{\theta} \,\, \text{non-dominated} \}$\;
        $(P, \, i) \gets (P \setminus F_i, \, i + 1)$\;
    }
    \textbf{return} $F_1, \dots, F_m$\;
}
\Hline{}
\Fn{\normalfont\texttt{selector\_eps\_net}($P$, $n$)}{
    $F_1, \dots, F_m \gets \texttt{non\_dom\_sorting}(P)$\;
    $\mathcal{C} \gets [ F_1\text{.pop}() ]$ \;
    \For{\normalfont $F = F_1, \dots, F_m$}{
        \While{\normalfont $F \neq \emptyset$}{
            $\vec{\theta} \gets \min_{\vec{\theta} \in F} \max_{\vec{\theta}' \in \mathcal{C}} \| \vec{y}_{\vec{\theta}} - \vec{y}_{\vec{\theta}'} \|_2 $\;
            $\mathcal{C}\text{.append}(\vec{\theta})$ \;
            $F \gets F \setminus \{\vec{\theta}\}$ \;
        }
    }
    \textbf{return} $\mathcal{C}_1, \dots, \mathcal{C}_n$ \;
}
\Hline{}
\Fn{\normalfont\texttt{selector\_nsga\_ii}($P$, $n$)}{
    $F_1, \dots, F_m \gets \texttt{non\_dom\_sorting}(P)$\;
    $\mathcal{C} \gets [ \, ]$ \;
    \For{\normalfont $F = F_1, \dots, F_m$}{
        $\mathcal{C} \gets \mathcal{C} + \texttt{crowding\_dist\_sorting}(F)$ \;
    }
    \textbf{return} $\mathcal{C}_1, \dots, \mathcal{C}_n$ \;
}
\end{algorithm}
\end{minipage}
\hfill
\begin{minipage}{0.46\textwidth}
\begin{algorithm}[H]\small
\caption{Multi-objective ASHA}
\label{algo:mo-asha}
\DontPrintSemicolon
\SetAlgoLined
\KwData{$R$, $r_0$, $s$, $\eta$ (default $\eta = 3$)}
\BlankLine
\Fn{\normalfont\texttt{mo\_asha}()}{
    \Repeat{\normalfont budget exhausted}{
        \For{\normalfont each free worker}{
            $(\vec{\theta}, k) \gets \texttt{get\_job}()$\;
            \texttt{run\_evaluation}($\vec{\theta}$, $r_0 \eta^{s + k}$)\;
        }
        \For{\normalfont completed job $(\vec{\theta}, k)$ with objective vector $\vec{y}_{\vec{\theta}}$}{
            update $\vec{\theta}$ in rung $k$ with $\vec{y}_{\vec{\theta}}$\;
        }
    }
}
\Hline{}
\Fn{\normalfont\texttt{get\_job}()}{
    \For{$k = \lfloor \log_\eta(R / r_0) \rfloor - s - 1, \dots, 1, 0$}{
        $\text{candidates} \gets \texttt{mo\_selector}(\text{rung} \, k, \left| \text{rung} \, k \right| / \eta)$\;
        $\text{promotable} \gets \{ \vec{\theta} \in \text{candidates}: \vec{\theta} \,\, \text{not promoted} \}$\;
        \If{\normalfont $\, | \text{promotable} | > 0$}{
            \textbf{return} (promotable[0], $k + 1$)\;
        }
    }
    draw random configuration $\vec{\theta}$ \;
    \textbf{return} $(\vec{\theta}, 0)$\;
}
\end{algorithm}
\end{minipage}
\end{figure*}
To mitigate these issues we propose to move away from scalarization. %

Instead, we make %
use of 
\textbf{non-dominating sorting} %
for budget allocation. In particular we adapt the NSGA-II~\cite{Deb2002:Fast} selection mechanism and the $\epsilon$-net~\cite{Salinas2021:Multi} (EpsNet) exploration strategy which ranks candidates in the same Pareto front by iteratively selecting the candidate with the largest euclidean distance from the set previously selected configurations. The pseudo-code for non-dominating sorting is presented in Algorithm~\ref{algo:nondominated}, as well as the structure of %
EpsNet and NSGA-II.  %
They both use a non-dominating sorting sub-routine, and they differ by their internal inner selection loop (please refer to the original NSGA-II paper \cite{Deb2002:Fast} for a complete definition of $\texttt{crowding\_dist\_sorting}$). In contrast to scalarization based techniques which assign a single value to each candidate \textit{independently} of other configurations, EpsNet and NSGA-II use information about the global geometry of the Pareto front approximation. They use non-dominating sorting to categorize evaluated candidates into multiple fronts and rank candidates in the individual fronts in a way that encourages a uniform coverage of the true Pareto front. By making decisions informed by the global geometry of the current Pareto front approximation these selections strategies encourage a more uniform exploration of the different objectives and are less impacted by objective values of different magnitudes.

We further replace the distributed synchronous scheduler used in Hyperband by the \textbf{distributed asynchronous scheduler} used in ASHA 
to improve wall-clock time. In each rung (i.e., a round of resources consumed), Hyperband waits until all configurations have exhausted their resource allocation. Especially when optimizing large neural networks where some configurations take much longer to evaluate than others this can be expensive because it causes idle workers. ASHA %
avoids this issue %
by assigning 
workers new configurations as soon as they terminate their current evaluation. It does so by operating on multiple rungs in parallel and by making its allocation decision based on partial information. 
This procedure has been shown to yield large wall-clock time improvements in the single objective setting. An overview of our proposed method is provided by Algorithm~\ref{algo:mo-asha}. It defines a family of MO-ASHA algorithms that can be equipped with candidate selection strategies. As concrete choices of selection strategies we experimentally evaluate the three scalarization schemes from Equation~\ref{eq:scalarizations} as well as the non-dominated sorting strategies NSGA-II and EpsNet.

%% file: text/background.tex
\section{Multi-objective Optimization Tasks}
\label{sec:background}

We consider three challenging MO hyperparameter optimization problems as detailed next.

\paragraph{Task 1: Neural Architecture Search (NAS)}
The goal of neural architecture search (NAS) is to automatically find network architectures where the search space may include number of convolutional layers, size of the convolutions or type of pooling. While initial works aims only at finding architectures with high prediction accuracy~\cite{zoph2017neural}, deploying such models (e.g., on the edge) also requires to consider %
other objectives such as cost or latency. %
We refer to~\cite{Benmeziane2021} for a survey on hardware-aware NAS. 
In particular, scalarization schemes have been considered in Computer Vision~\cite{Tan2018,Wu2019}, where the objectives were mapped to a single real value via a fixed set of weights trying to achieve a good trade-off. 
Having to fix the weights a priori is impractical and restrictive as
each specific scalarization can only capture restricted Pareto shapes~\cite{Golovin2020}. Other approaches such as~\cite{Elsken2019} propose to %
tailor EA to NAS by modeling the nature of %
the multiple competing objectives explicitly. %
Considering these recent trends, we conducted experiments on the NAS-201 benchmark ~\cite{Dong2020:Nasbench201} which contains tabular evaluations of accuracy and latency for 15625 architectures on 3 datasets.

\paragraph{Task 2: Algorithmic Fairness}
Algorithmic fairness %
aims to train accurate models subject to fairness criteria.
Prior work has proposed several fairness criteria, whose suitability depends on the application domain, and it has been shown that criteria,commonly referred to acceptance-based and error-based, can be in conflict with each other~\cite{friedler2016possibility,fairness2018verma,kleinberg2018inherent}. %

Let $Y$ be the true label, $\hat{Y}$ the predicted label, and $S$ a binary sensitive attribute %
(i.e., the population is formed by two subgroups). We consider the following criteria:
\begin{itemize}
\item \textit{Statistical Parity} (SP) requires equal Positive Rates (PR) across subgroups: %
$P( \hat{Y} = 1 | S = 0 ) = P( \hat{Y} = 1 | S = 1 )$,
\item \textit{Equal Opportunity} (EO) requires equal True Positive Rates (TPR) across subgroups: %
$P( \hat{Y} = 1 | Y = 1, S = 0 ) = P( \hat{Y} = 1 | Y = 1, S = 1 )$;
\item \textit{Equalized Odds} (EOdd) requires equality of False Positive Rates (FPR) in addition to EO. 
\end{itemize}
The extension to more than two sensitive features values and more than two classes is straightforward.

In general, algorithmic fairness methods can be divided into three families: (i) {\it post-processing} to modify a pre-trained model to increase the fairness of its outcomes~\cite{feldman2015certifying,hardt2016equality,pleiss2017fairness}; (ii) {\it in-processing} to enforce fairness constraints during training~\cite{agarwal2018reductions,donini2018empirical,zafar2017fairness,zafar2019fairness}; and (iii) {\it pre-processing} to modify the data representation and then apply standard machine learning algorithms~\cite{calmon2017optimized,zemel2013learning}.
Most of %
approaches provide solutions for only a single fairness metric. For example, \cite{feldman2015certifying} is limited to demographic parity definition of fairness. In contrast to in-processing methods, which
are often dependent on the model class (e.g.,~\cite{zafar2017fairness,Zhang2018,donini2018empirical}) and hence have limited extensibility, using MO-HPO allows us to consider any fairness constraint and %
while being agnostic to the specific model as it operates {\it only} on its hyperparameters. %

\paragraph{Task 3: Training Language Models}
Language modeling assign a probability value to a given input text sequence $\mathbf{t} = (t_1, t_2, \ldots, t_T)$ by modeling individual token probabilities as 
$ p(\mathbf{t}) = \prod_{i=1}^{T} p(t_i | t_1, t_2, \ldots, t_{i-1})$. 
The most popular evaluation metric for language models (LM) is perplexity over a held-out set. A high perplexity indicates that the LM is ``surprised'' to see the input sequences. As a result, a low perplexity is a desirable property. %
However, perplexity is shown to not always correlate with other desirable metrics such as word error rate~\cite{Chen1998:Evaluation}. Word error rate measures the accuracy of the LM in predicting the next word by comparing the word that is predicted to be most likely by the LM with the ground truth next word in the sequence. Moreover, while larger LMs can lead to lower perplexity, it may also result in higher prediction time due to increased network size. In this work, we consider Transformer-based LMs that not only minimize perplexity, but also lead to low prediction time and low word error rate.

%% file: text/experiments.tex
\section{Experiments}
\label{sec:experiments}
We evaluate our approach on the three different HPO tasks discussed in Section~\ref{sec:background}: (i) Neural architecture search using NAS-201 benchmark~\cite{Dong2020:Nasbench201}, (ii) algorithmic fairness training MLP models on Adult dataset~\cite{Kohavi1996:Scaling} and (iii) Transformer language models optimization on Wikitext2~\cite{Merity2016:Pointer} corpus. Hyperparameter spaces are provided by Tables~\ref{tab:nas parameter},~\ref{tab:mlp parameter} and~\ref{tab:wikitext2 parameter} in Appendix~\ref{app:experimental_setup}.
We compare MO-ASHA implementations employing 3 different scalarization techniques (RW, ParEGO, Golovin), 2 globally informed selection strategies (NSGA-II, EpsNet) and random search (RS) baseline. Following~\cite{Schmucker2020:Multi} each scalarization method is equipped with 100 random vectors. As ASHA hyperparameters we chose reduction factor $\eta = 3$, minimum resource budget $r_0 = 1$ and maximum budget $R = 200$ for NAS-201 and Adult and $R = 81$ for Wikitext2.
For our evaluations, we normalize all objectives values with quantiles using the empirical CDF which scales objective ranges to the $[0, 1]$ interval making our comparisons robust to monotonic transformations of the objectives \cite{Binois2020}. This allows us to compute the dominated hypervolume with respect to worst case reference point $\Vec{1} \in \mathbb{R}^n$. For Adult and Wikitext2 a reference Pareto front approximation $\mathcal{A}$ is formed by accumulating the evaluations from all runs and epochs. The tabular nature NAS-201 allows us to determine the exact Pareto front.

\paragraph{NAS-201 benchmark dataset}
\begin{figure}[t]
    \centering
        \includegraphics[width=0.49\textwidth]{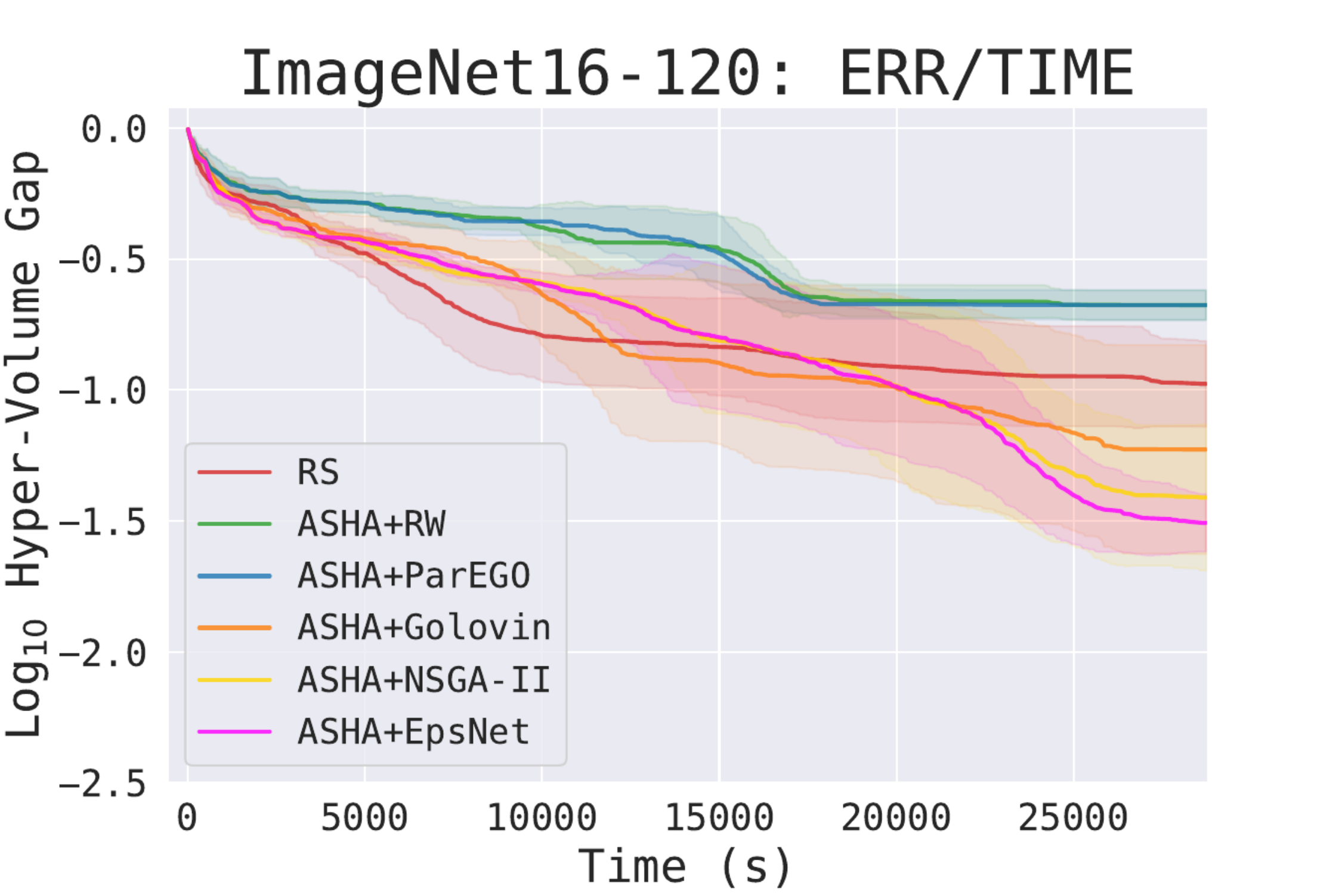}
        \includegraphics[width=0.49\textwidth]{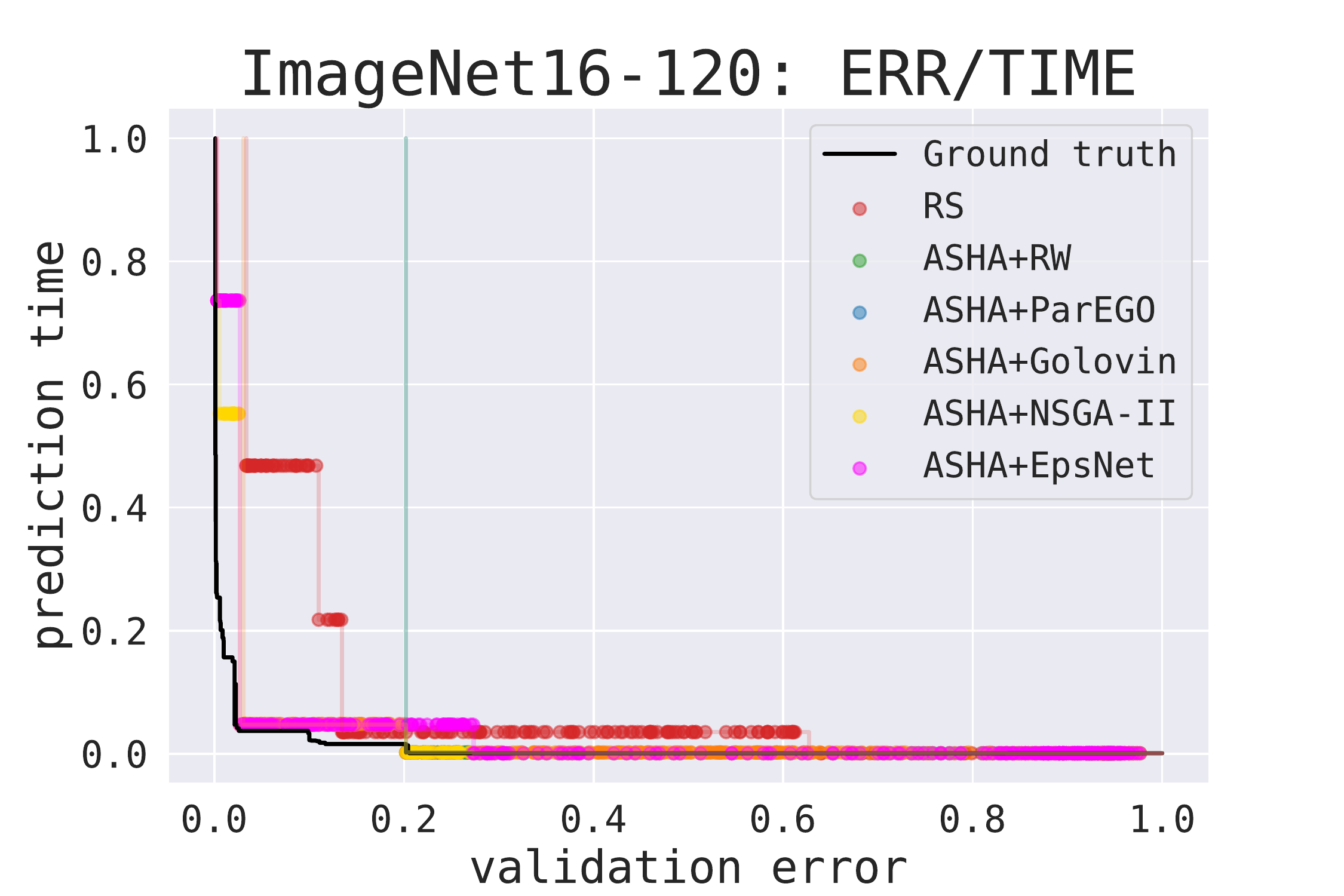}
    \caption{\small Left: Difference in dominated hypervolume of the Pareto front approximations for the different methods with respect to the real Pareto front approximation over time for ImageNet16-120 dataset. The objectives of interest are two: error (ERR) and prediction time (TIME). Right: Pareto fronts found by the different methods for the same dataset, compared to the ground truth Pareto front (i.e., the best Pareto front that can be obtained from NAS-201 network architectures).}
    \label{fig:nas201-ImageNet16}
\end{figure}

\begin{figure}[t]
    \centering
        \includegraphics[width=0.95\textwidth]{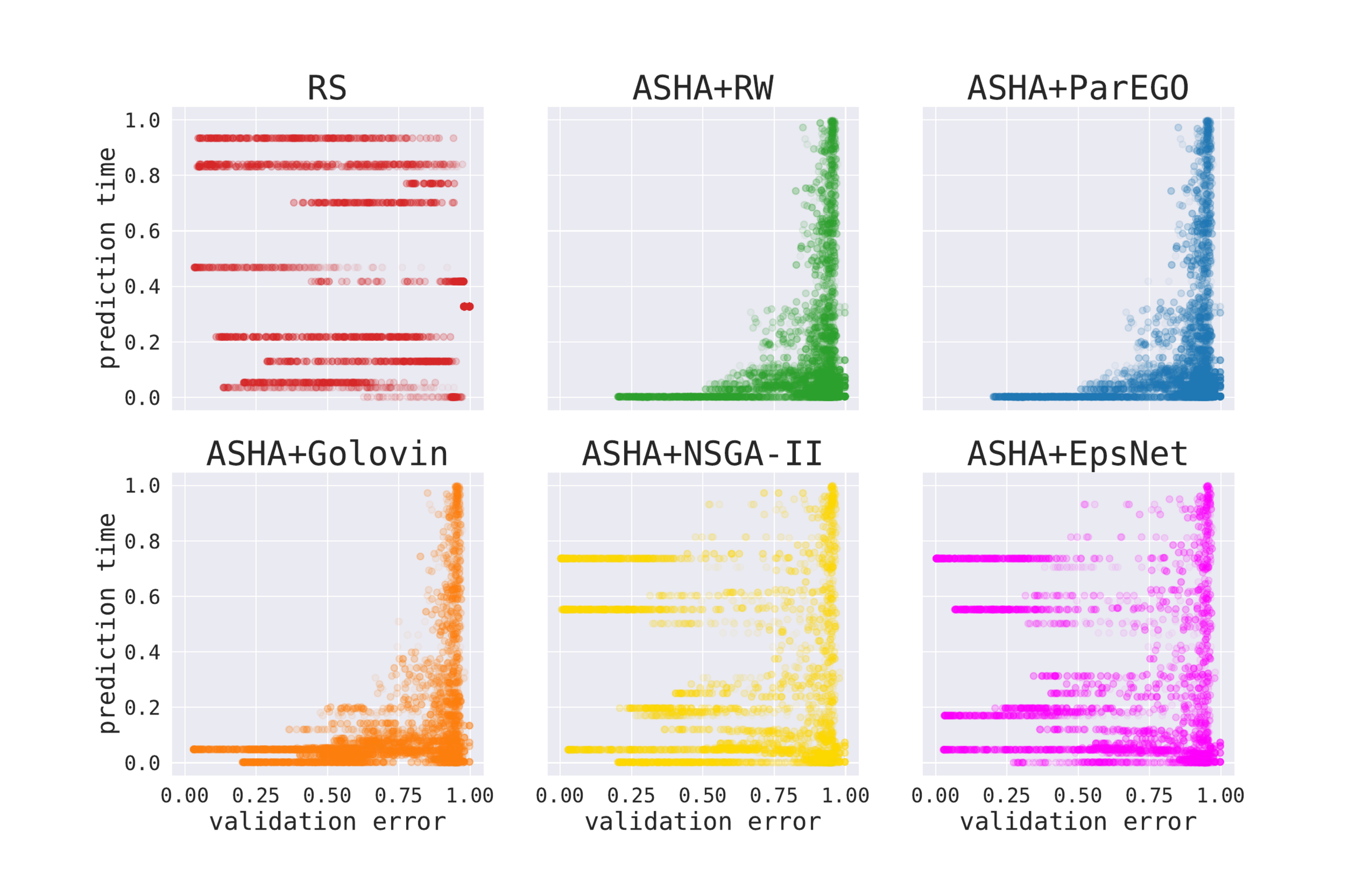}
    \caption{\small Objective values visualizing how different methods explore the NAS-201 ImageNet16-120 architecture space over time. Each model evaluation is represented by a single dot. Darker color indicates that the model has been evaluated later in time during the search.}
    \label{fig:exploration-imagenet16}
\end{figure}
In the first experiment we search for fast and accurate Res-Net architectures using the tabular NAS-201 benchmark dataset. NAS-201 describes a 6-dimensional search space and provides evaluations of 15625 different neural network architectures on 3 datasets. We perform 10 runs using 4 parallel workers on AWS ml.c4.xlarge instances. We set a maximum run time of 3 hours for Cifar-10, 5 hours for Cifar-100 and 8 hours for ImageNet16-120 datasets. We optimize over two objectives: the validation error (ERR) and the prediction or inference time (TIME).

The left side of Figure~\ref{fig:nas201-ImageNet16} visualizes the difference in dominated hypervolume of the approximations to the real ImageNet16-120 Pareto front over time. The plot shows the performance of the different methods, highlighting that the group of methods relying on scalarization techniques (RW, ParEGO, Golovin) obtains worse performance than the methods using geometric information (NSGA-II, EpsNet). Especially, EpsNet obtains the best performance at the end of the time budget. In this task, RS performs sufficiently well, outperforming RW and ParEGO, and being very competitive in the first phases of the exploration, due to the full training it performs for each choice of hyperparameters.

The right side of Figure~\ref{fig:nas201-ImageNet16} shows the Pareto fronts recovered by the different methods, as well as the ground truth front (i.e., the Pareto front of all possible NAS-201 networks). It is interesting to note how scalarization techniques tend to penalize one objective heavier than the other, focusing on models with very low prediction time, and avoiding the area of the search space with slower more accurate models. This shows the advantage of globally informed algorithms being able to explore the search space more uniformly. In fact, globally informed techniques are more robust towards objectives of different magnitude. %
Experimental results for Cifar-10 and Cifar-100 datasets are provided in Appendix \ref{app:additional_experiments}. Cifar-100 confirms the insights we presented for ImageNet16-120. Cifar-10 is a rather simple problem and all MO-ASHA variants perform roughly the same.

Finally, another important aspect to analyse is how the different algorithms explore the search space over time. Similar dominated hypervolume can be achieved with different exploration strategies. We visualize the models evaluated by the different methods (considering each epoch as an evaluation) over time in Figure \ref{fig:exploration-imagenet16} to see which parts of the search space they focus on.
Each dot represents an evaluation and darker colors indicates evaluation at a later point in time. We can observe that RW, ParEGO and Golovin tend to focus heavily on finding models with low prediction times and fail to explore slower models that yield the best accuracy. Random search does not show this bias, but suffers from inefficient resource allocation. NSGA-II and EpsNet have a more balanced exploration scheme that on one side explores faster models with good accuracy, but also considers slower candidates which leads to the discovery of very accurate models.

\paragraph{Fair model for Adult dataset}
\begin{figure}
    \centering
        \includegraphics[width=0.49\textwidth]{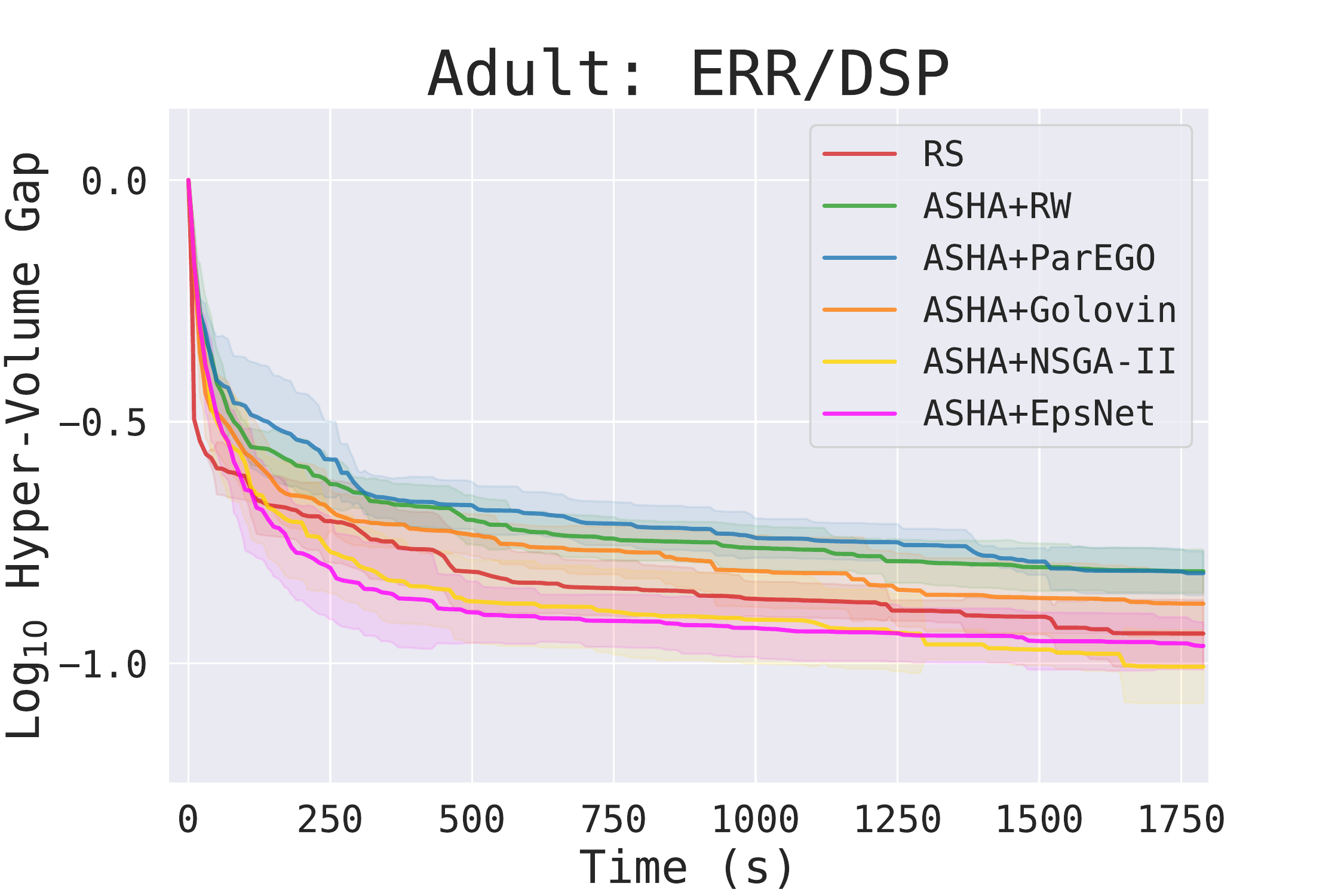}
        \includegraphics[width=0.49\textwidth]{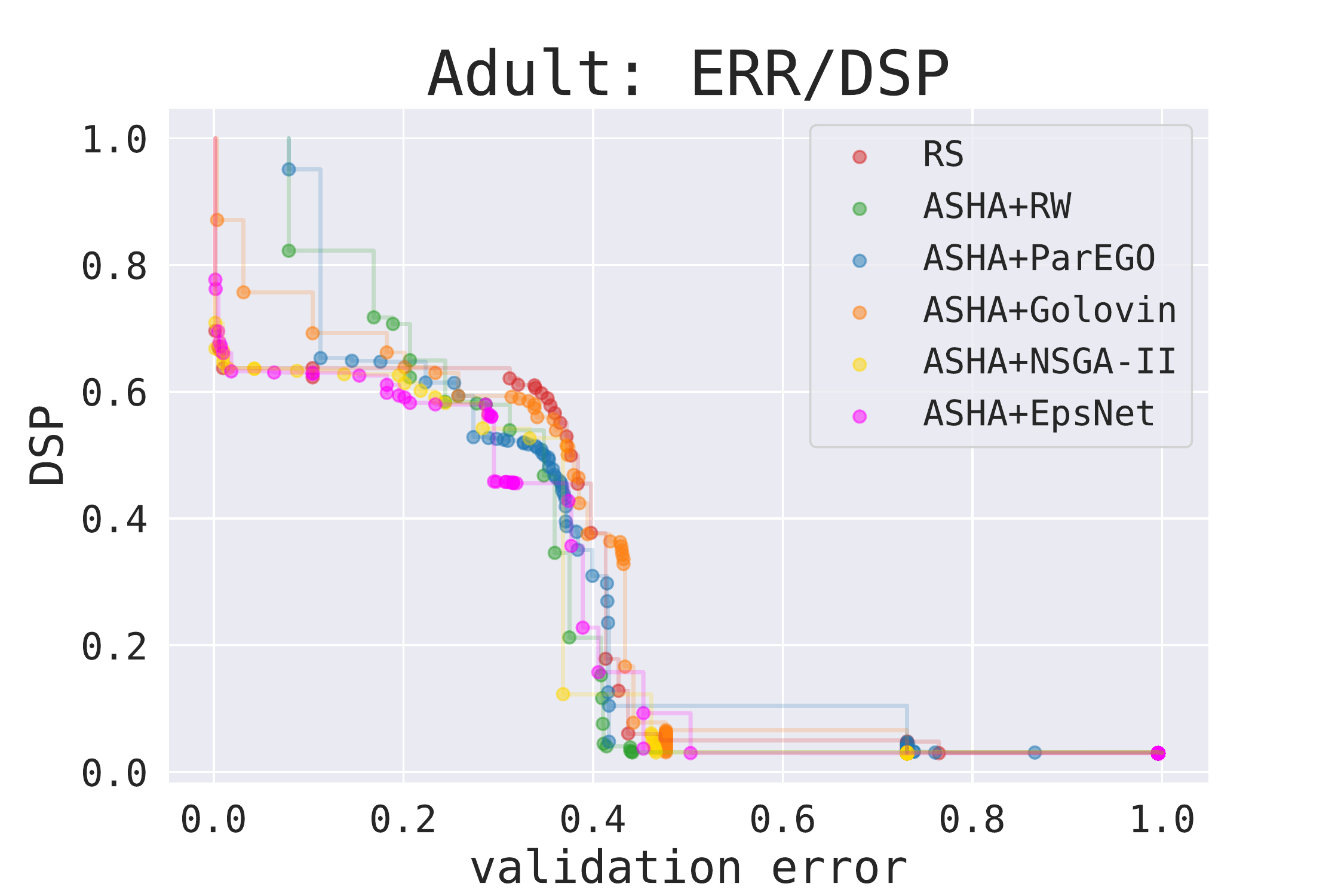}
    \caption{\small Left: Difference in dominated hypervolume of the Pareto front approximations for the different methods with respect to the combined front approximation over time on Adult with two objectives: ERR and DSP. Right: Pareto fronts found by the different methods on the same dataset.}
    \label{fig:adult}
\end{figure}
\begin{table}
\small
\centering
\caption{\small Average validation error (and standard deviation) over 10 repetitions for the best fair models for different methods. We use the fairness constraint: DSP $\le 0.1$.}
\begin{tabular}{l|c||l|c}
\hline
Fairness Method & Validation error & MO Method & Validation error \\
\hline
FERM &  0.164 $\pm$ 0.010 & RS & 0.166 $\pm$ 0.005 \\
Zafar &  0.187 $\pm$ 0.001 & ASHA+RW & 0.166 $\pm$ 0.001  \\
Adversarial & 0.237 $\pm$ 0.001 &ASHA+ParEGO & 0.167 $\pm$ 0.003  \\
FERM pre-processed &  0.228 $\pm$ 0.013 & ASHA+Golovin & 0.164 $\pm$ 0.002  \\
SMOTE &  0.178 $\pm$ 0.005 & ASHA+NSGA-II & 0.165 $\pm$ 0.003 \\
CBO MLP & 0.167 $\pm$ 0.017 & ASHA+EpsNet & 0.166 $\pm$ 0.004  \\
\hline
\end{tabular}
\label{tab:fairsota}
\end{table}
For the second set of experiments we selected the Adult~\cite{Kohavi1996:Scaling} dataset and search for MLP models that are both accurate and fair with respect to a sensitive attribute. The Adult dataset is based on census data, and contains binary gender as sensitive attribute. The binary classification task is to predict if the annual income of an individual exceeds 50K USD. This dataset is commonly used to compare the performance of different algorithmic fairness techniques. The search space of considered MLP models is 10 dimensional. Because the dataset is rather small and training a model until convergence is fast we reduced the maximum run time to 30 minutes. We perform 10 runs using 4 parallel workers on AWS ml.c4.xlarge instances and use a 70\%/30\% random split to form training and validation set.

We train MLP models, optimizing for low validation error (ERR) and different fairness objectives: Statistical Parity (SP), Equal Opportunity (EO) and Equalized Odds. Specifically, we use the violation of the fairness constraint as a measure of unfairness (i.e., objective). In the case of EO, we introduce {\it difference in equal opportunity} (DEO), that is the absolute value of the difference of the TPRs between the two sub-groups. And, analogously, we define {\it difference in statistical parity} (DSP) and {\it difference in false positive rates} (DFP). 

The left side of Figure~\ref{fig:adult} visualizes the difference in dominated hypervolume of the individual front approximations to combined approximation $\mathcal{A}$ over time. The experiments confirm the observations made in the NAS setting. NSGA-II and EpsNet outperform the other methods, with Golovin performing slightly better than RW and ParEGO performing worst. Interestingly, in this task, RS proves to be a competitive baseline at the very beginning (due to complete model training for each parameter choice) and at the end of the 30 minute runs. Aligning with these observations, Appendix \ref{app:additional_experiments} shows results for 3- and 4-objective settings combining various fairness criteria. The right side of Figure~\ref{fig:adult} visualizes the Pareto front approximations recovered by the different methods. Unlike in the NAS-201 experiments we do not have access to the ground truth Pareto front.  Interestingly, this Pareto front is concave which explains the poor performance of linear scalarization methods as they can only recover convex fronts.

It is important to note how using a MO approach allows us to be agnostic towards the underlying machine learning algorithm and combination of objectives (here fairness constraints). This is crucial considering that many the state-of-the-art algorithmic fairness methods can only accommodate specific models or specific fairness definitions. In order to compare the models obtained using the MO approaches to state-of-the-art algorithmic fairness methods we followed the same experimental setting proposed in \cite{perrone2020fair} and \cite{Schmucker2020:Multi}. We select a single measure of fairness, i.e. DSP, and find the model with highest accuracy such that the DSP value is less or equal $0.1$. This is a common scenario in algorithmic fairness, where $0.1$ DSP is often considered acceptable.

In Table \ref{tab:fairsota}, we report the results of this experiment, comparing MO algorithms versus the following algorithmic fairness methods: Zafar \cite{zafar2017fairness}, Adversarial debiasing \cite{Zhang2018}, Fair Empirical Risk Minimization (FERM and FERM pre-processed)~\cite{donini2018empirical}, constrained BO approach (CBO) for fair models~\cite{perrone2020fair}, and SMOTE~\cite{chawla2002smote}.\footnote{Code for Zafar from \url{https://github.com/mbilalzafar/fair-classification}; code for Adversarial Debiasing from \url{https://github.com/IBM/AIF360}; code for FERM from \url{https://github.com/jmikko/fair_ERM}; code for CBO can be found in AutoGluon \url{https://auto.gluon.ai}.} These results, show that the MO approach is very competitive compared to model- and constraint-specific methods (FERM, Zafar, Adversarial), and also without knowing \emph{a priori} the threshold of our fairness constraint (as instead needed for CBO).

\paragraph{Language Models for Wikitext2 dataset}
\begin{figure}[t]
    \centering
        \includegraphics[width=0.49\textwidth]{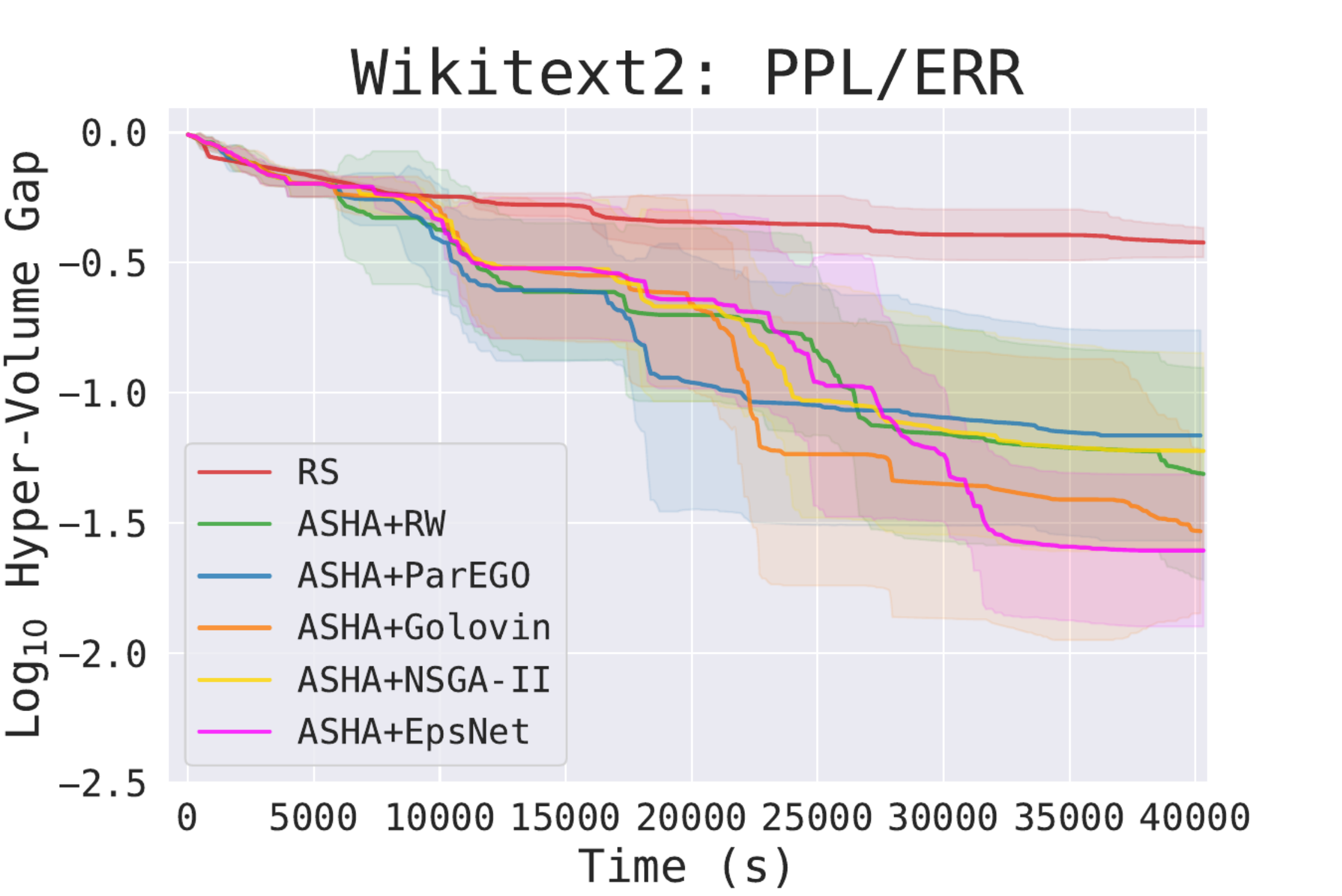}
        \includegraphics[width=0.49\textwidth]{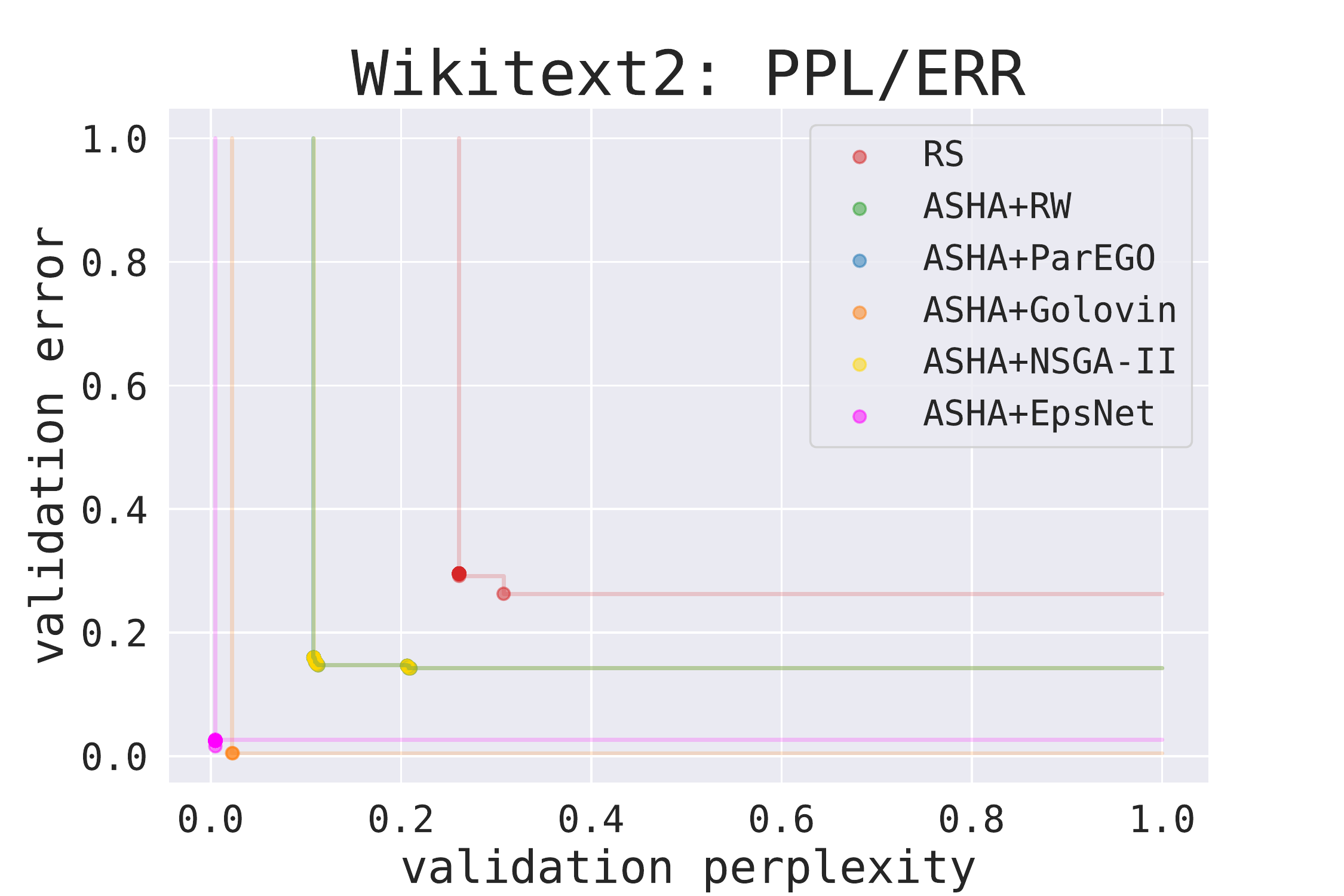}
    \caption{Left: Difference in dominated hypervolume of the Pareto front approximations for the different methods with respect to the combined front approximation over time on Wikitext2 with two objectives: PPL and ERR. Right: Pareto fronts found by the different methods on the same dataset.}
    \label{fig:wikitext2_ppl_err}
\end{figure}
In the last set of experiments we optimize Transformer based language models on the Wikitext2 corpus for perplexity (PPL), word error rate (ERR) and inference time (TIME). The search space described in Appendix~\ref{app:experimental_setup} captures 6 dimensions. We perform 5 runs using a single worker on AWS ml.g4dn.xlarge instances and set the maximum run time to 12h. This experiment is computationally demanding which forced us to perform fewer runs. In total we used 720 GPU hours / 1 GPU month. 

The left side of Figure~\ref{fig:wikitext2_ppl_err} visualizes the difference in dominated hypervolume of the Pareto front approximations to combined Pareto front approximation $\mathcal{A}$. All methods except random search perform roughly the same during the first 3h of the experiment. ParEGO and Golovin show good performance towards the 7h mark. Towards the end of the time budget, EpsNet obtains the best Pareto front approximations. The right side of Figure~\ref{fig:wikitext2_ppl_err} shows the Pareto fronts recovered by the different methods. Unlike in the two prior experiments perplexity and word error rate are not very conflicting with each other and the main problem is to decide which models to allocate the computational budget too. Taking a look at the performance of random search it is easy to see that naive random sampling does not lead to satisfactory performance. Figure~\ref{fig:exploration-wikitext2_ppl_err} in Appendix~\ref{app:additional_experiments} visualizes how the different exploration strategies explore the search space over time. We observe correlation between perplexity and word error rate objectives. Compared to the experiments on the other datasets we are only able to evaluate a rather small number of configurations. Random search spends a lot of computational resources on configurations that turn out to be unsuitable. Overall, the figure underlines the point that the main challenge in this task is efficient resource allocation. Given a larger computational budget it would be interesting to see how the different approaches compare to each other at a point of saturation (i.e. how long does it take each method until it does not discover further useful candidates).

%% file: text/conclusion.tex
\section{Conclusion}
\label{sec:conclusion}
In this paper we presented multiple extensions of ASHA to MO in the context of HPO. We evaluated the performance of several methods across three distinct tasks: neural architectural search, algorithmic fairness and language modeling, where the goal of optimizing multiple objectives rises naturally. Our experiments showed that MO ASHA enables MO HPO at scale. Our empirical analysis showed the best performance when MO ASHA is used in synergy with EpsNet selection mechanism to utilize the global geometry of the Pareto front approximation. Our proposal outperforms multi-fidelity HPO based on multi-objective scalarization with respect to the wall-clock time performances by a large margin. All the implemented algorithms will be open-sourced in AutoGluon~\cite{agtabular}.

%% file: text/appendix.tex
\section{Experimental Setup}
\label{app:experimental_setup}
Adding to Section~\ref{sec:experiments}, we provide additional details about the experimental setup. Descriptions of the hyperparameter spaces used for the NAS-201, Adult, and Wikitext2 experiments are provided by Tables~\ref{tab:nas parameter},~\ref{tab:mlp parameter} and~\ref{tab:wikitext2 parameter} respectively. For parameters that capture values ranging across multiple orders of magnitude we apply a logarithmic scaling. The NAS-201 and Adult experiments were all performed using AWS ml.c4.xlarge instances providing 4 cores of an Intel Xeon E5-2666 v3 CPU as well as 7.5GB of RAM. For the Wikitext2 experiments we used AWS ml.g4dn.xlarge instances providing 4 cores of a second generation Intel Xeon Cascade Lake CPU, 16GB of RAM as well as one NVIDIA T4 Tensor Core GPU. For the Adult experiment we used Scikit-learn~\cite{scikit-learn} (version 0.24) and for the Wikitext2 experiment we used PyTorch~\cite{Paszke2019:PyTorch} (version 1.8.1).
\begin{table}[H]
    \center
    \caption{NAS-201, ResNet search space}
    \begin{tabular}{lrr}
        \hline
         Parameter &    Type &    Domain  \\
        \hline
         $x_0$ &    categorical &    \{skip\_connect, conv\_1x1, conv\_3x3, pool\_3x3, none \} \\
        \hline
         $x_1$ &    categorical &    \{skip\_connect, conv\_1x1, conv\_3x3, pool\_3x3, none \} \\
        \hline
         $x_2$ &    categorical &    \{skip\_connect, conv\_1x1, conv\_3x3, pool\_3x3, none \} \\
        \hline
         $x_3$ &    categorical &    \{skip\_connect, conv\_1x1, conv\_3x3, pool\_3x3, none \} \\
        \hline
         $x_4$ &    categorical &    \{skip\_connect, conv\_1x1, conv\_3x3, pool\_3x3, none \} \\
        \hline
         $x_5$ &    categorical &    \{skip\_connect, conv\_1x1, conv\_3x3, pool\_3x3, none \} \\
        \hline
    \end{tabular}
    \label{tab:nas parameter}
\end{table}
\begin{table}[H]
    \center
    \caption{Adult, Sklearn MLP search space}
    \begin{tabular}{lrrr}
        \hline
        Parameter &    Type &    Domain &     Scaling \\
        \hline
        n\_layers &    integer &    $\{1, 2, 3, 4\}$ &    linear\\
        \hline
        layer\_1 &    integer &    $\{2, \dots, 32\}$ &    linear\\
        \hline
        layer\_2 &    integer &    $\{2, \dots, 32\}$ &    linear\\
        \hline
        layer\_3 &    integer &    $\{2, \dots, 32\}$ &    linear\\
        \hline
        layer\_4 &    integer &    $\{2, \dots, 32\}$ &    linear\\
        \hline
        alpha &    real &    $[10^{-6}, \dots, 10^{-1}]$ &    logarithmic\\
        \hline
        learning\_rate\_init &    real &    $[10^{-6}, \dots, 10^{-2}]$ &    logarithmic\\
        \hline
        beta\_1 &    real &    $[0.001, 0.99]$ &    logarithmic\\
        \hline
        beta\_2 &    real &    $[0.001, 0.99]$ &    logarithmic\\
        \hline        
        tol &    real &    $[10^{-5}, 10^{-2}]$ &    logarithmic\\
        \hline
    \end{tabular}
    \label{tab:mlp parameter}
\end{table}
\begin{table}[H]
    \center
    \caption{Wikitext2, Transformer search space}
    \begin{tabular}{lrrr}
        \hline
        Parameter &    Type &    Domain &     Scaling \\
        \hline
        lr &    real &    $[1, 50]$ &    logarithmic\\
        \hline
        dropout &    real &    $[0, 0.99]$ &    linear\\
        \hline
        batch\_size &    integer &    $\{8, \dots, 256\}$ &    linear\\
        \hline
        clip &    real &    $[0.1, 2]$ &    linear\\
        \hline
        lr\_factor &    integer &    $\{1, \dots, 100\}$ &    logarithmic\\
        \hline
        emsize &    integer &    $\{32, \dots, 1024\}$ &    linear\\
        \hline
    \end{tabular}
    \label{tab:wikitext2 parameter}
\end{table}

\newpage

\section{Additional Experimental Results}
\label{app:additional_experiments}
\begin{figure}[H]
    \centering
        \includegraphics[width=0.49\textwidth]{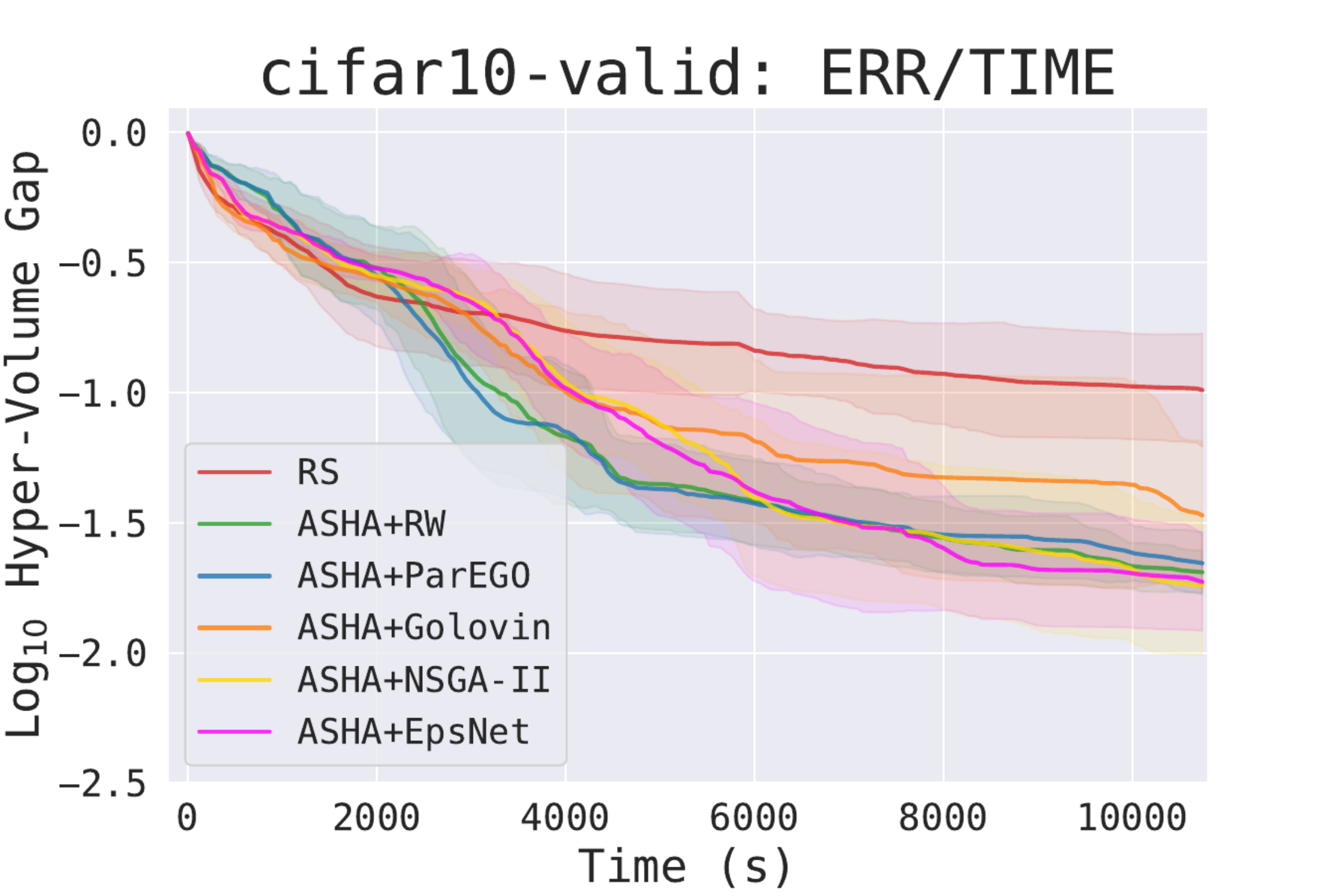}
        \includegraphics[width=0.49\textwidth]{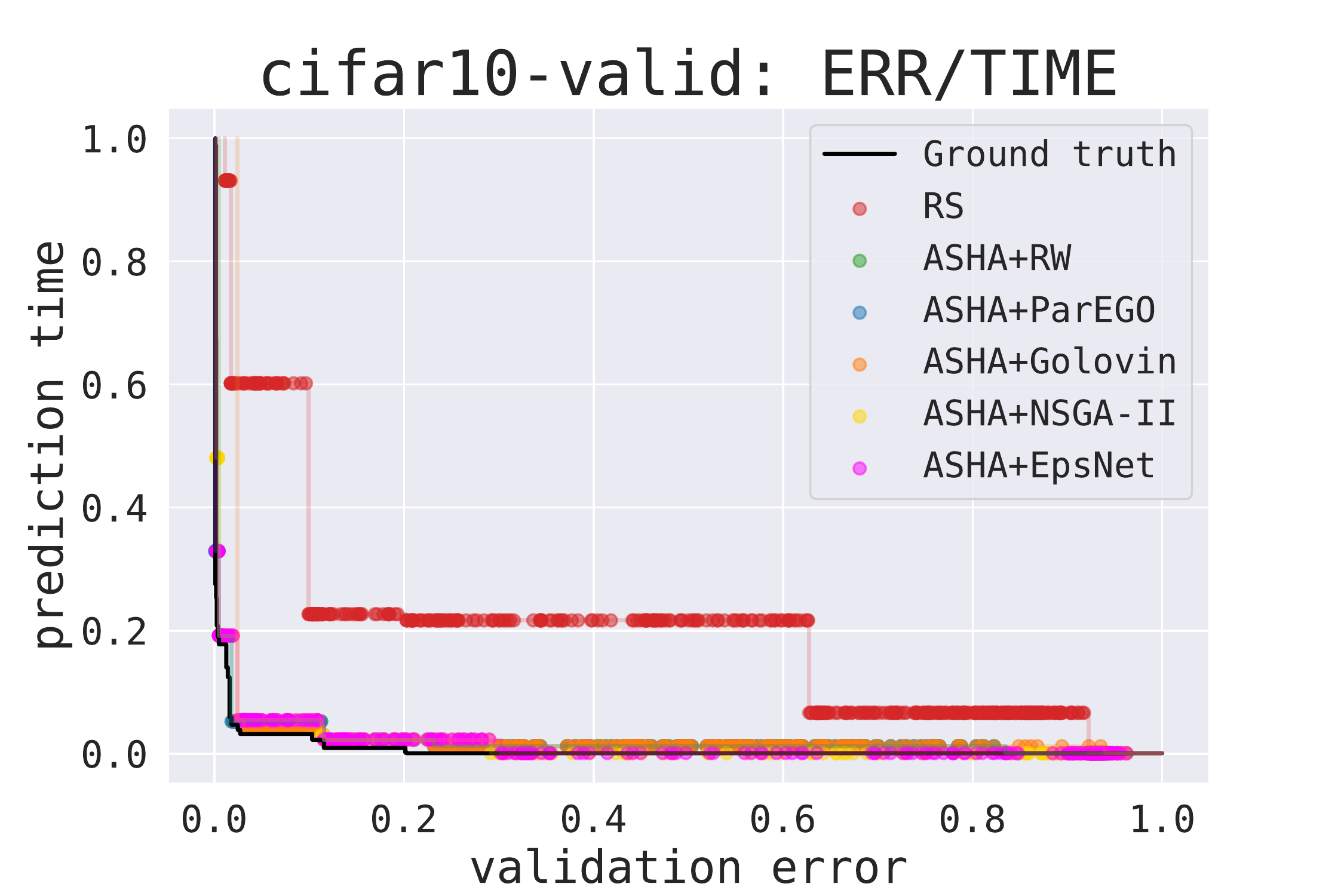}
    \caption{Left: Difference in dominated hypervolume of the Pareto front approximations for the different methods with respect to the optimal Pareto front approximation over time on NAS-201 Cifar-10 dataset with two objectives: ERR and TIME. Right: Pareto fronts found by the different methods on the same dataset.}
    \label{fig:nas201-cifar10}
\end{figure}

\begin{figure}[H]
    \centering
        \includegraphics[width=0.95\textwidth]{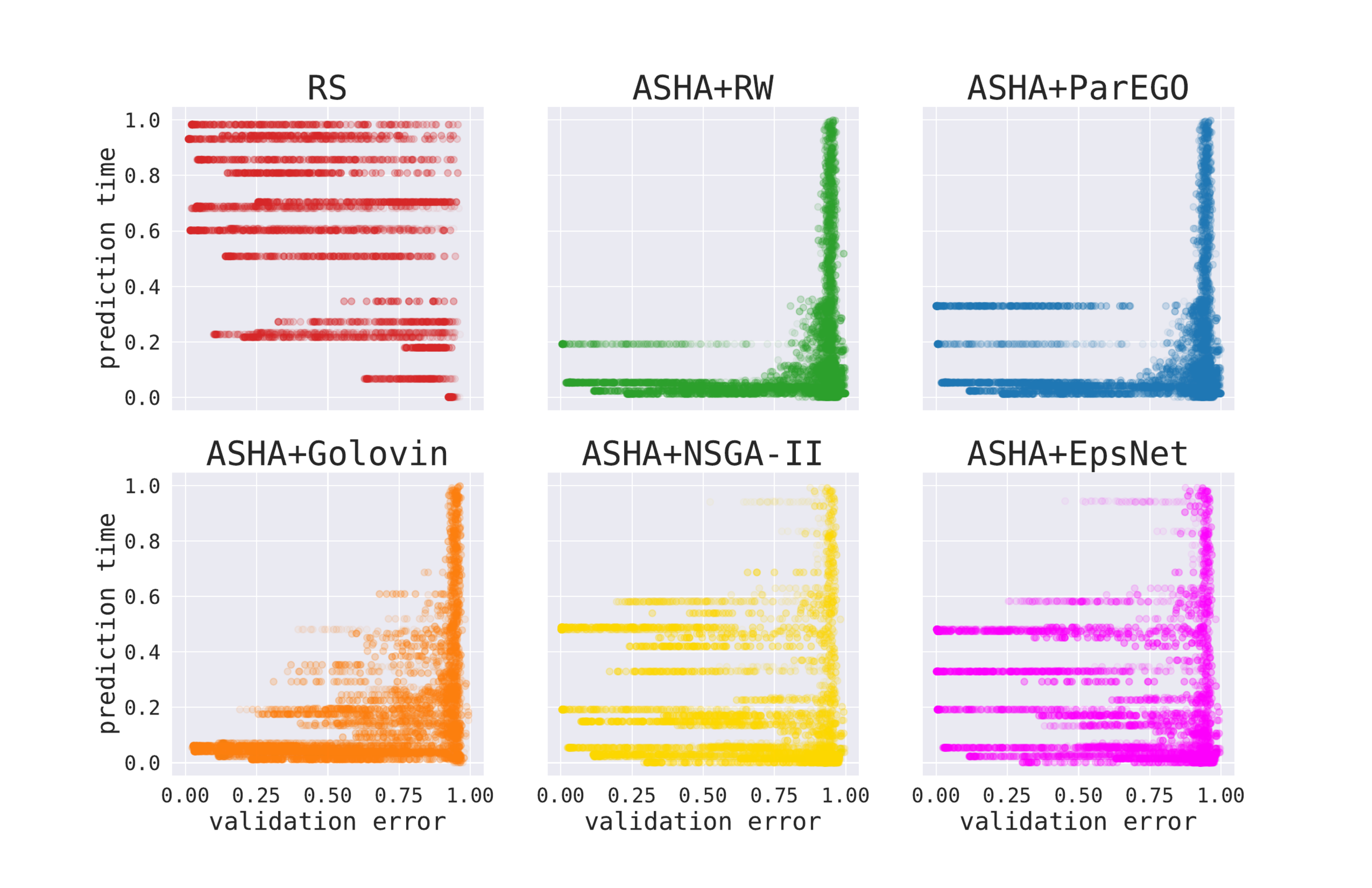}
    \caption{Objective values visualizing how different methods explore the NAS-201 Cifar-10 architecture space over time. Each model evaluation is represented by a single dot. Darker colors indicate that the model has been evaluated later in time during the search.}    \label{fig:exploration-cifar10}
\end{figure}

\begin{figure}[H]
    \centering
        \includegraphics[width=0.49\textwidth]{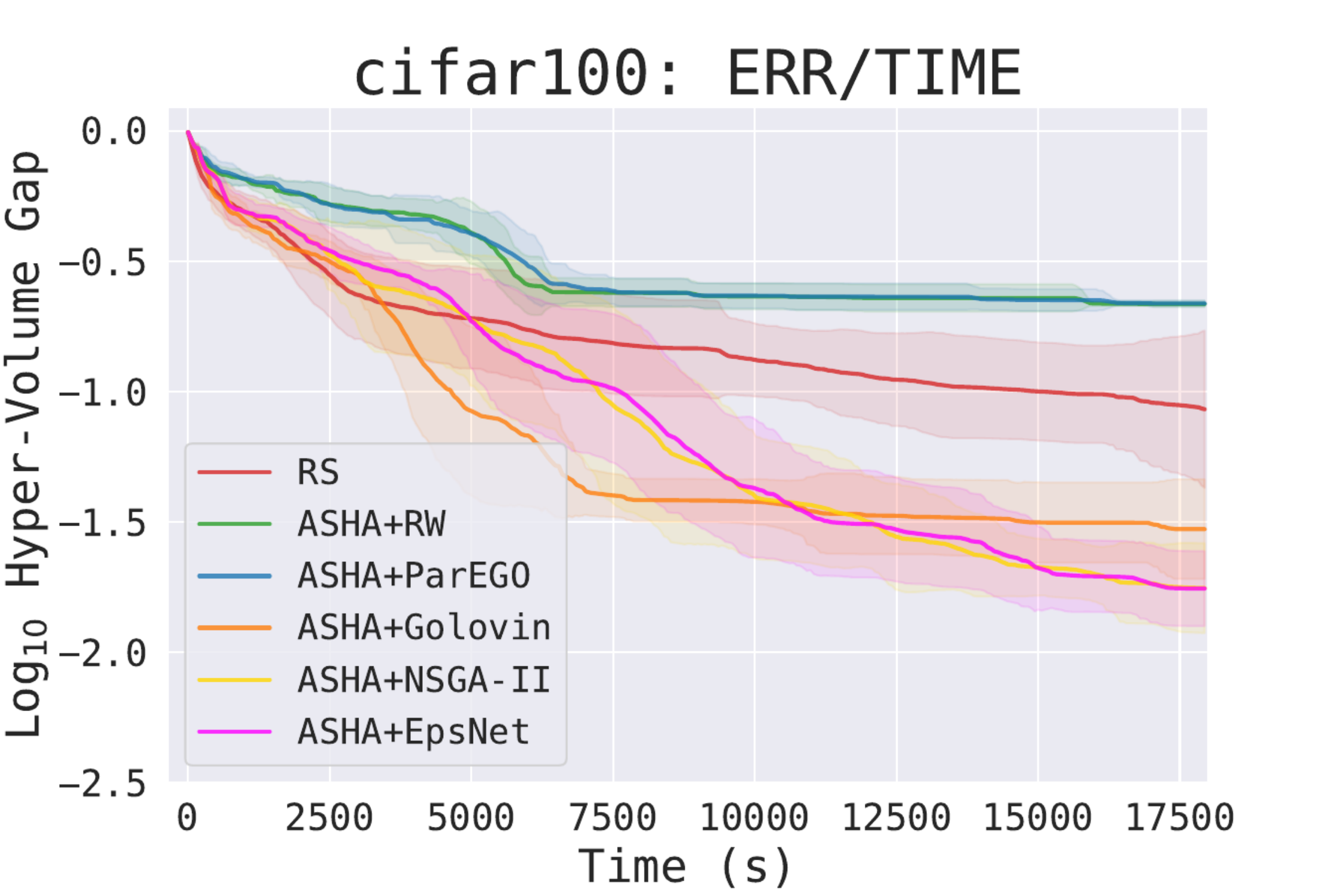}
        \includegraphics[width=0.49\textwidth]{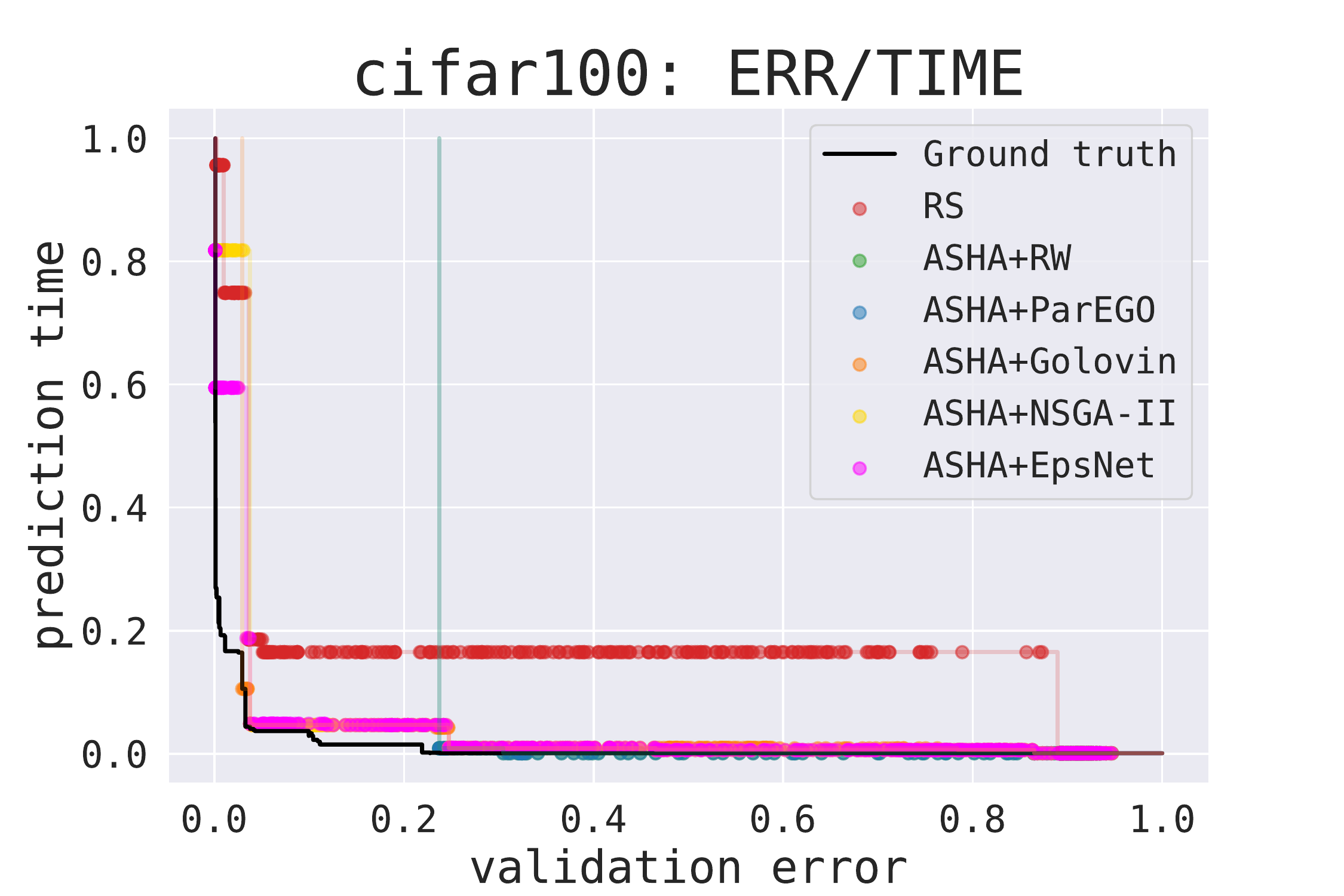}
    \caption{Left: Difference in dominated hypervolume of the Pareto front approximations for the different methods with respect to the optimal Pareto front approximation over time on NAS-201 Cifar-100 dataset with two objectives: ERR and TIME. Right: Pareto fronts found by the different methods on the same dataset.}
    \label{fig:nas201-cifar100}
\end{figure}

\begin{figure}[H]
    \centering
        \includegraphics[width=0.95\textwidth]{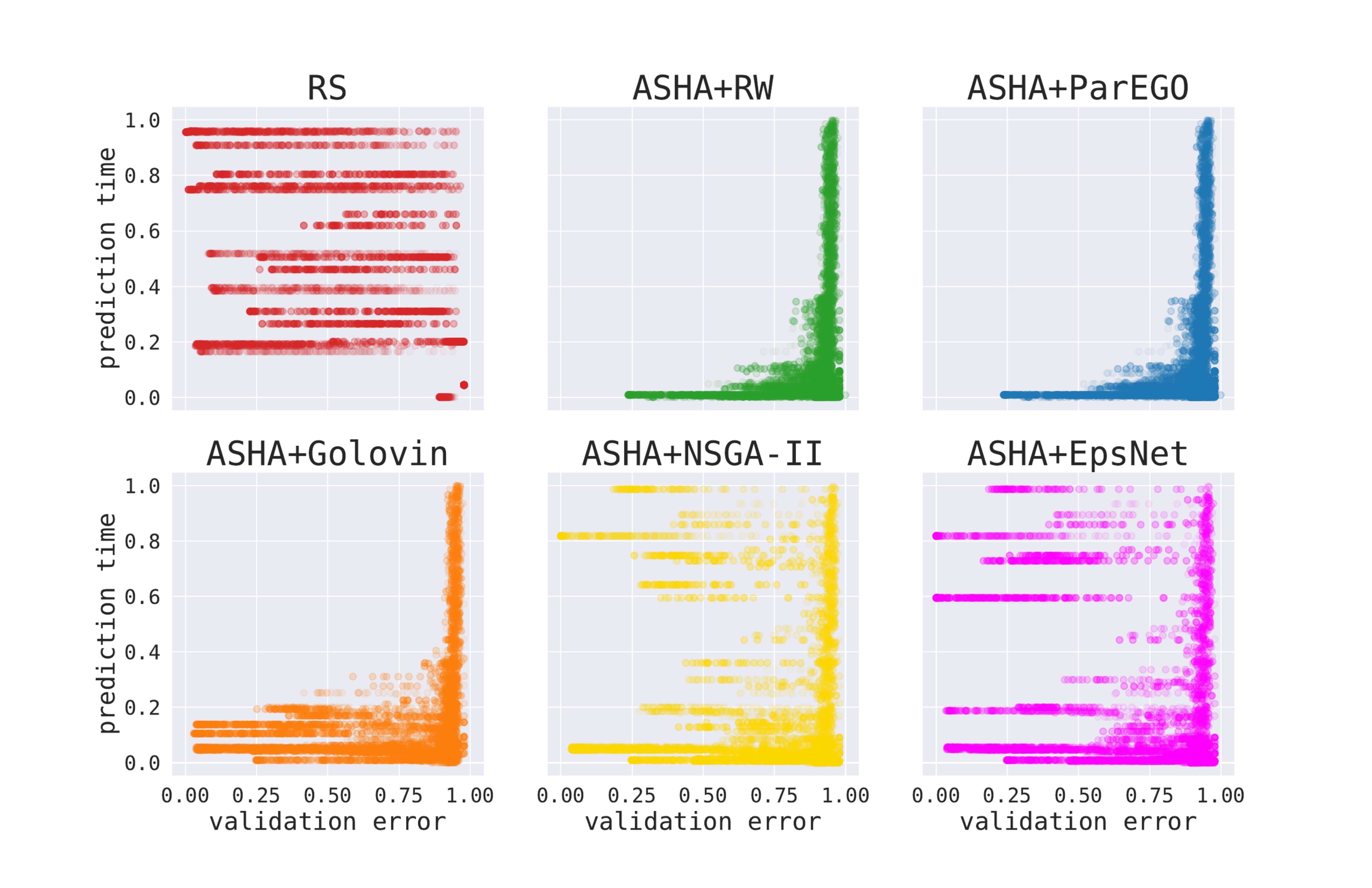}
    \caption{Objective values visualizing how different methods explore the NAS-201 Cifar-100 architecture space over time. Each model evaluation is represented by a single dot. Darker colors indicate that the model has been evaluated later in time during the search.}    \label{fig:exploration-cifar100}
\end{figure}

\begin{figure}[t]
    \centering
        \includegraphics[width=0.49\textwidth]{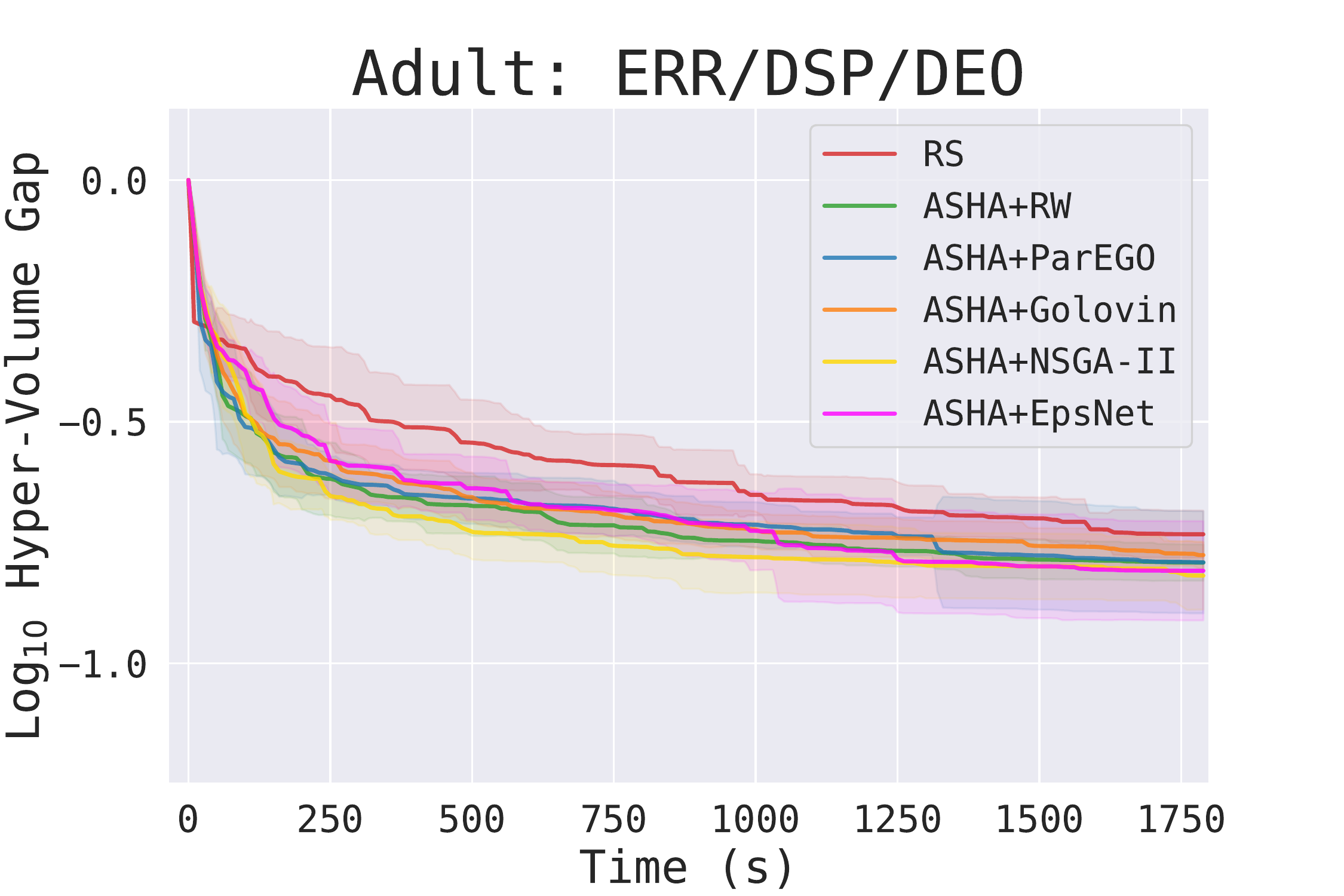}
        \includegraphics[width=0.49\textwidth]{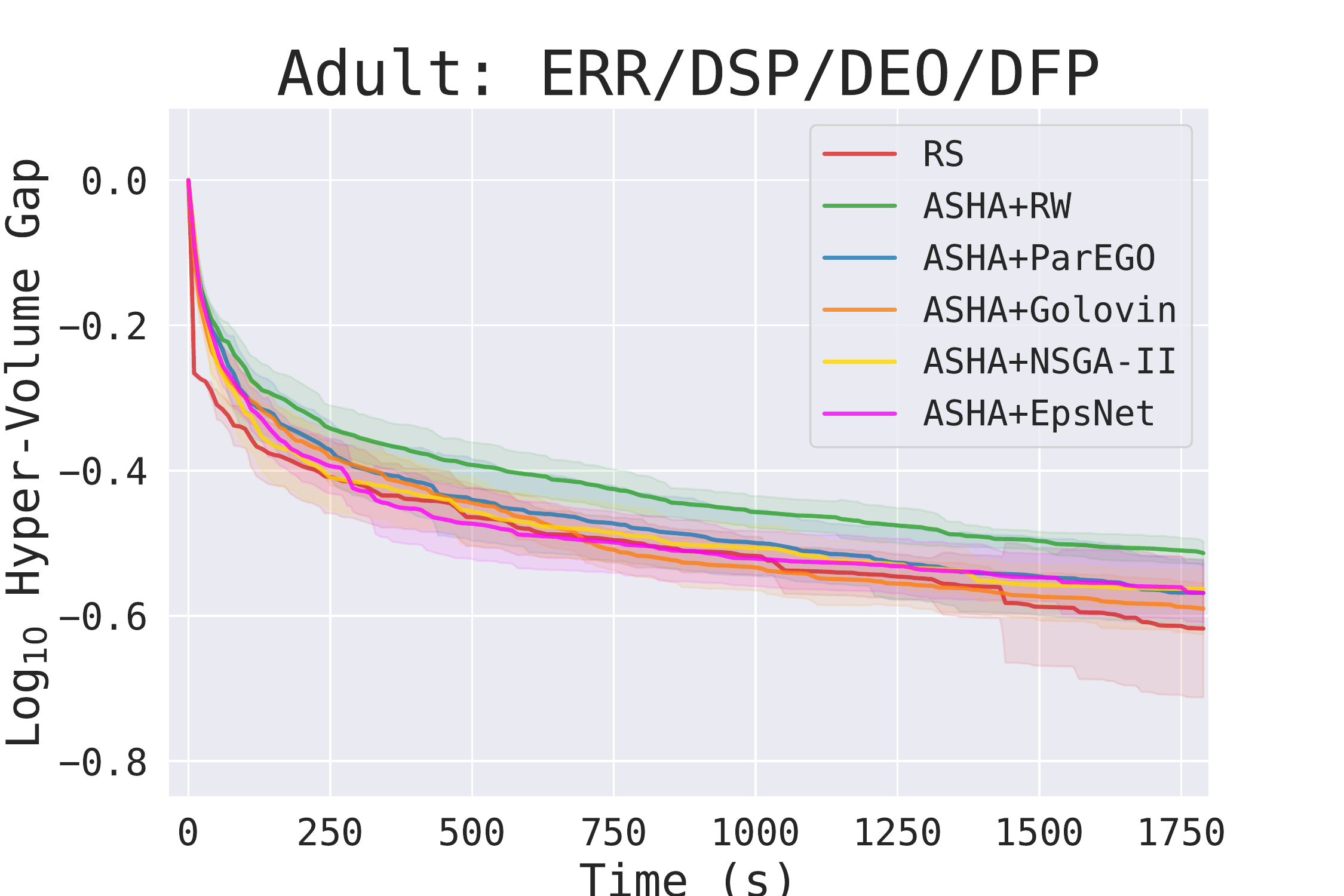}
    \caption{Left: Difference in dominated hypervolume of the Pareto front approximations for the different methods with respect to the combined front approximation over time on Adult with three objectives: ERR, DSP and DEO. Right: Analog with four objectives: ERR, DSP, DEO and DFP.}
    \label{fig:multi_adult}
\end{figure}

\begin{figure}[t]
    \centering
        \includegraphics[width=0.95\textwidth]{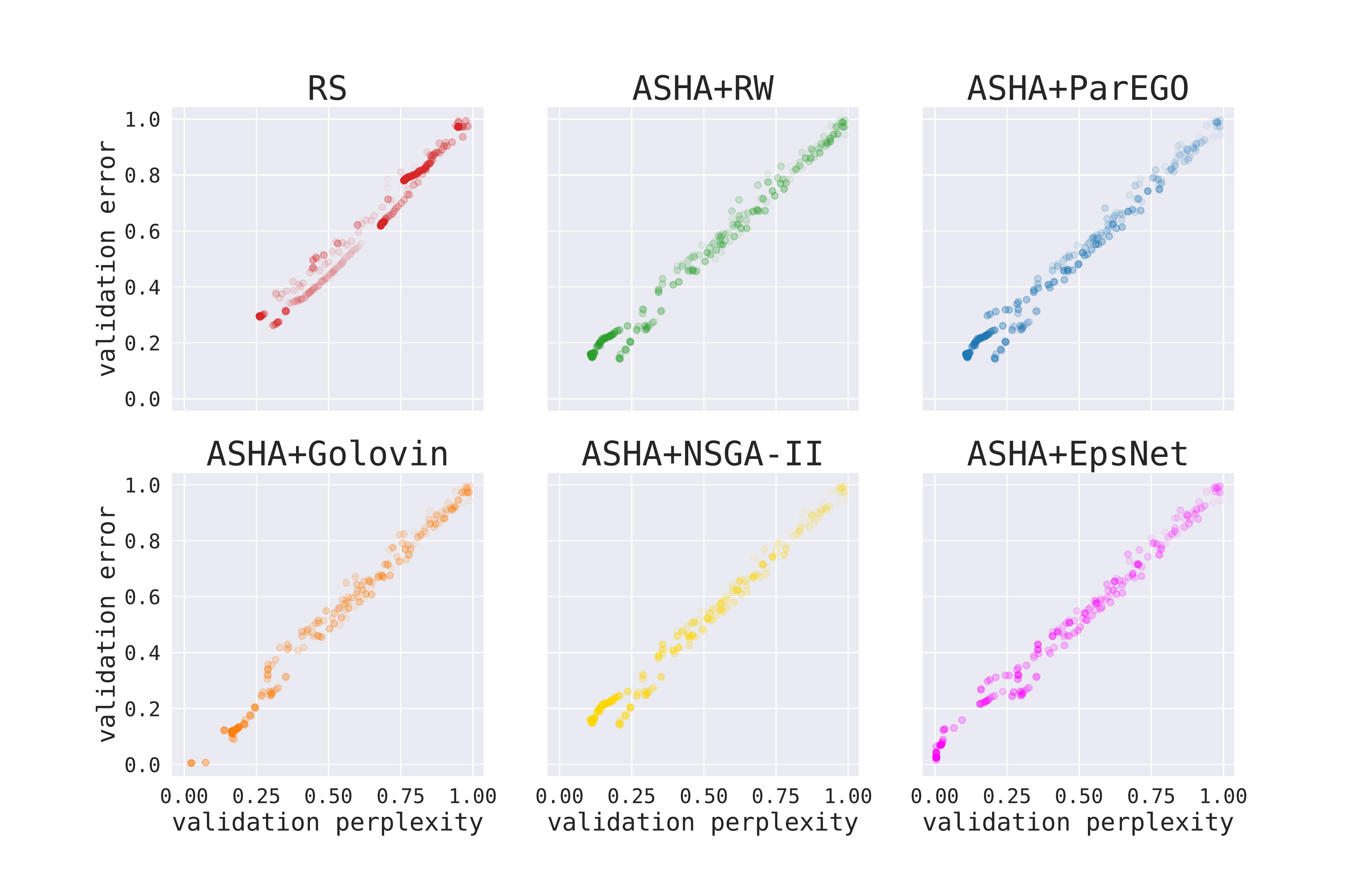}
    \caption{Objective values for the exploration of the search space by the different techniques on Wikitext2 dataset under PPL and ERR objectives. Each model evaluation is represented by a single dot, and darker colors indicate that the model has been evaluated later in time during the search.}
    \label{fig:exploration-wikitext2_ppl_err}
\end{figure}

\begin{figure}[H]
    \centering
        \includegraphics[width=0.49\textwidth]{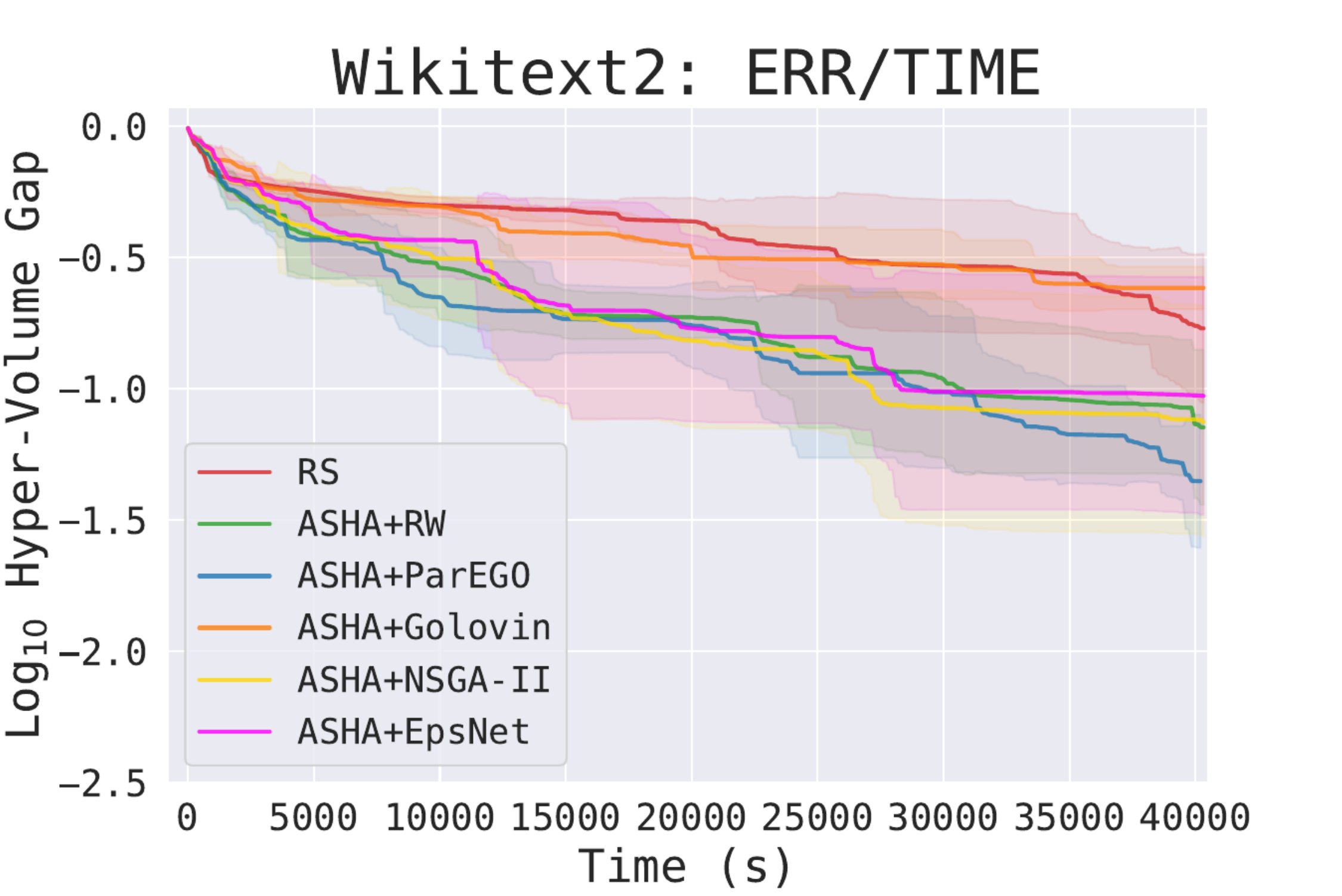}
        \includegraphics[width=0.49\textwidth]{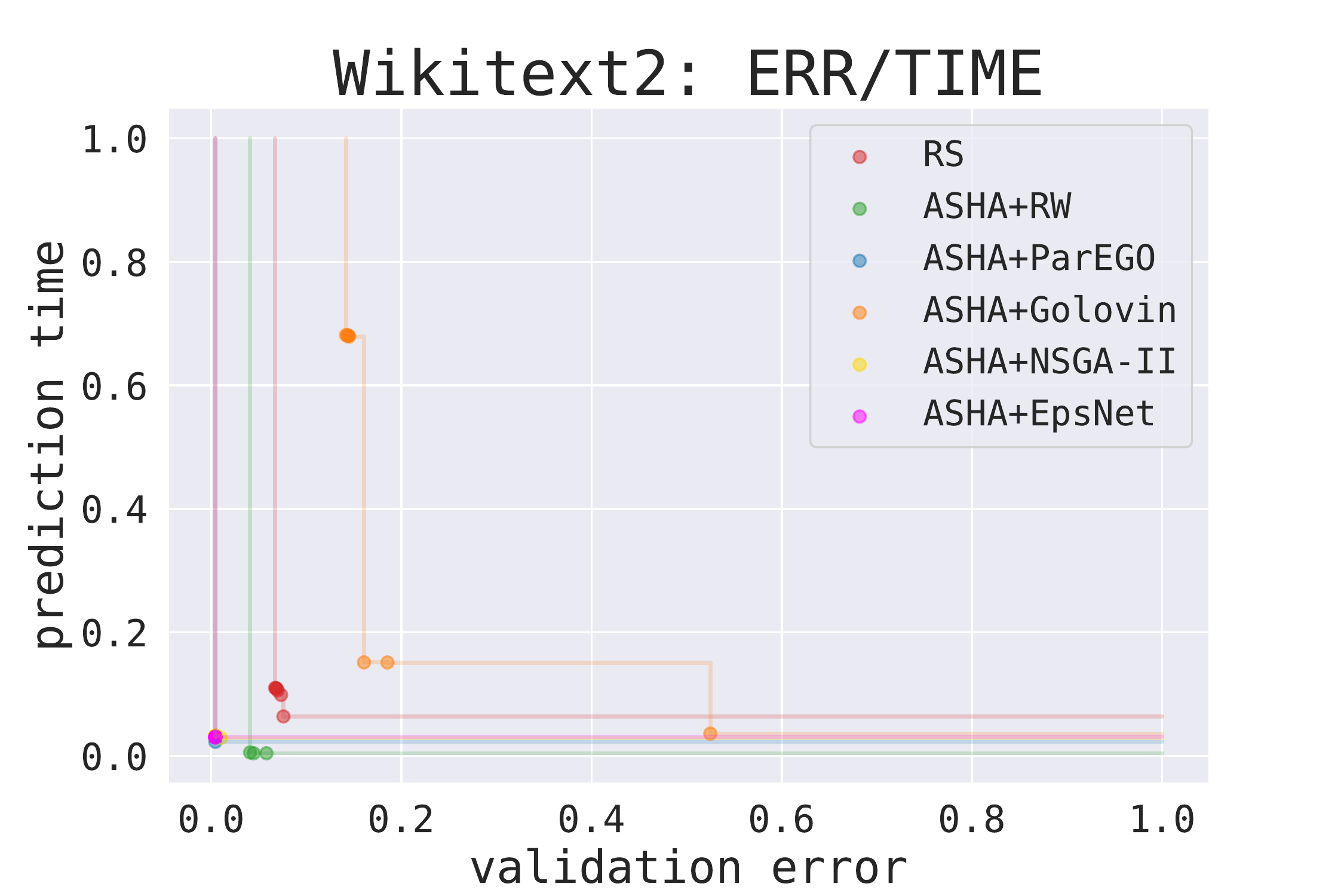}
    \caption{Left: Difference in dominated hypervolume of the Pareto front approximations for the different methods with respect to the combined front approximation over time on Wikitext2 dataset with two objectives: ERR and TIME. Right: Pareto fronts found by the different methods on the same dataset.}
    \label{fig:wikitext2_err_time}
\end{figure}

\begin{figure}[H]
    \centering
        \includegraphics[width=0.95\textwidth]{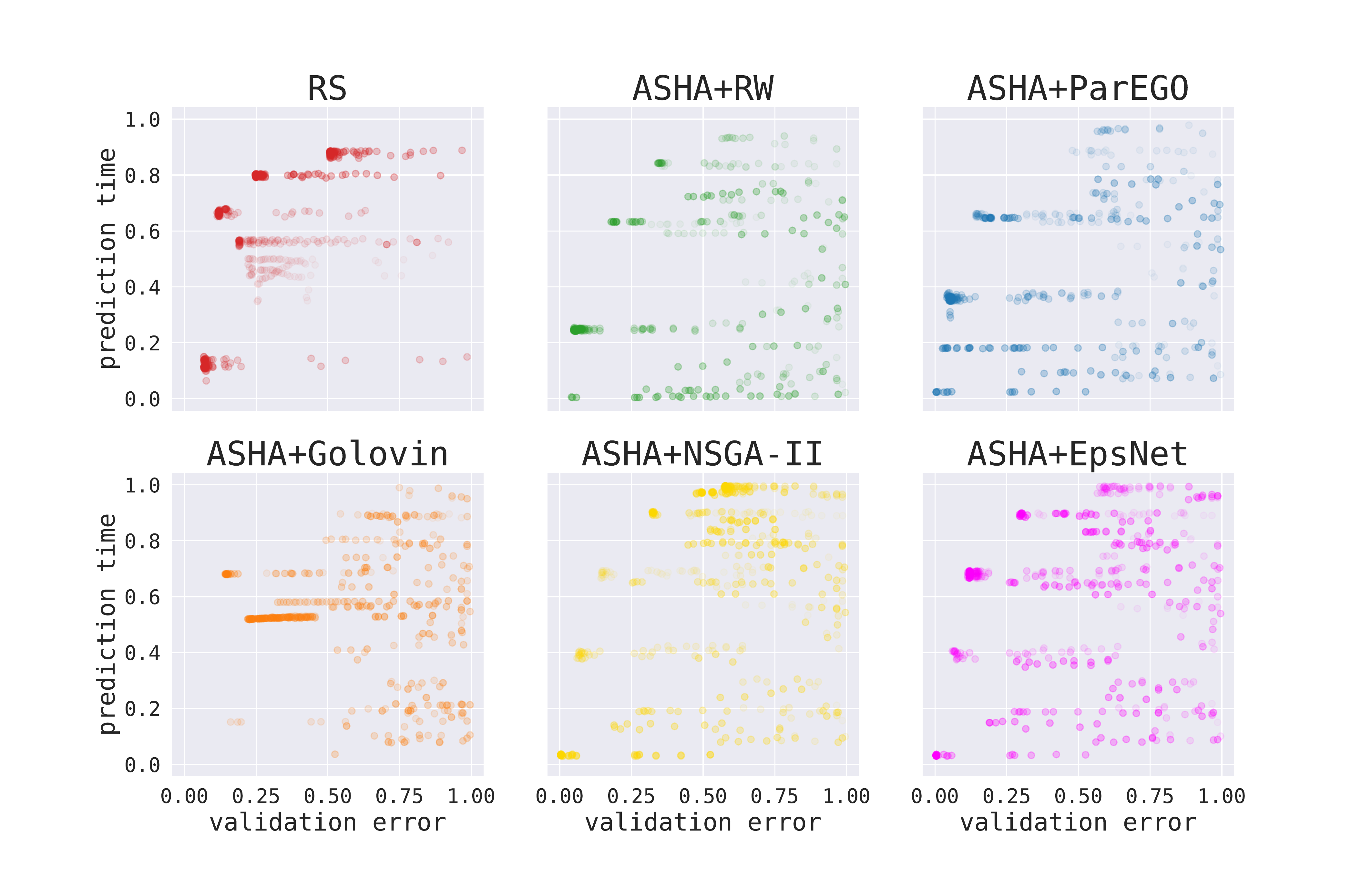}
    \caption{Objective values for the exploration of the search space for the different techniques on Wikitext2 dataset under ERR and TIME objectives. Each model evaluated is represented by a single dot, and darker colors indicate that the model has been evaluated later in time during the search.}
    \label{fig:exploration-wikitext2_err_time}
\end{figure}

Adding to Section~\ref{sec:experiments}, we visualize further experimental results. Figure~\ref{fig:nas201-cifar10} shows the hypervolume dominated by Pareto front approximations recovered by the different MO-ASHA methods and random search over time for the NAS-201 Cifar-10 dataset. Corresponding to that, Figure~\ref{fig:exploration-cifar10} visualizes how the different methods explore the search space over time. Analogously, Figure~\ref{fig:nas201-cifar100} and~\ref{fig:exploration-cifar100} show dominated hypervolume and exploration over time for the NAS-201 Cifar-100 dataset. The left and right side of Figure~\ref{fig:multi_adult} show the dominated hypervolume over time for a 3 objective (ERR, DSP, DEO) and a 4 objective (ERR, DSP, DEO, DFP) fair machine learning problem on the Adult dataset respectively. Adding to Figure~\ref{fig:wikitext2_ppl_err} in the main body, Figure~\ref{fig:exploration-wikitext2_ppl_err} indicates which parts of the search space the methods explore over time. Finally, Figure~\ref{fig:wikitext2_err_time} shows the dominated hypervolume over time for a language model task on the Wikitext2 experiment under word error rate (ERR) and prediction cost (TIME) objectives and correspondingly Figure~\ref{fig:exploration-wikitext2_err_time} visualizes the exploration of the search space over time.




\newpage

\section{Comparison with Multi-objective Bayesian Optimization}
\label{app:mbo_comparison}

\begin{figure}[H]
    \centering
        \includegraphics[width=0.49\textwidth]{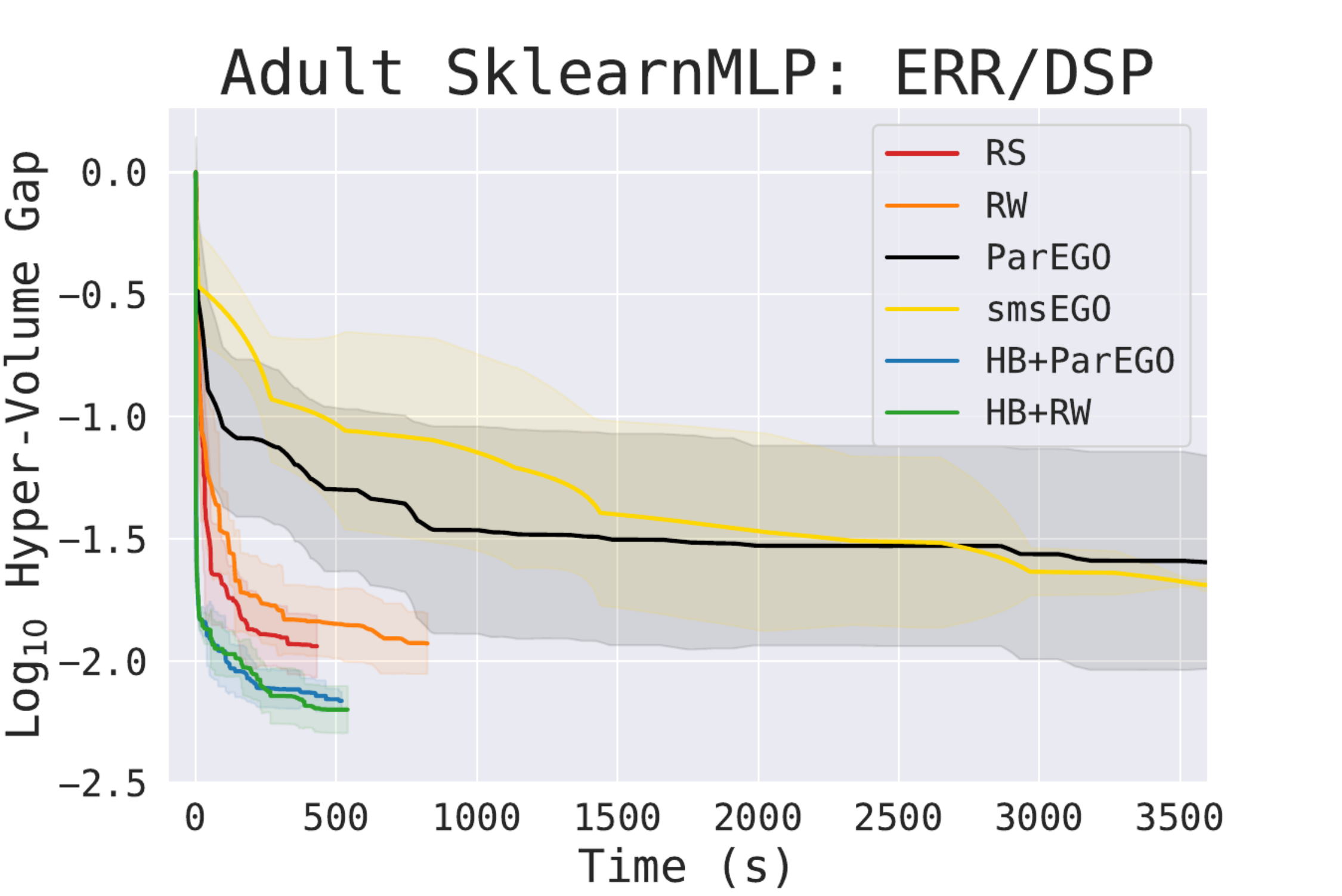}
        \includegraphics[width=0.49\textwidth]{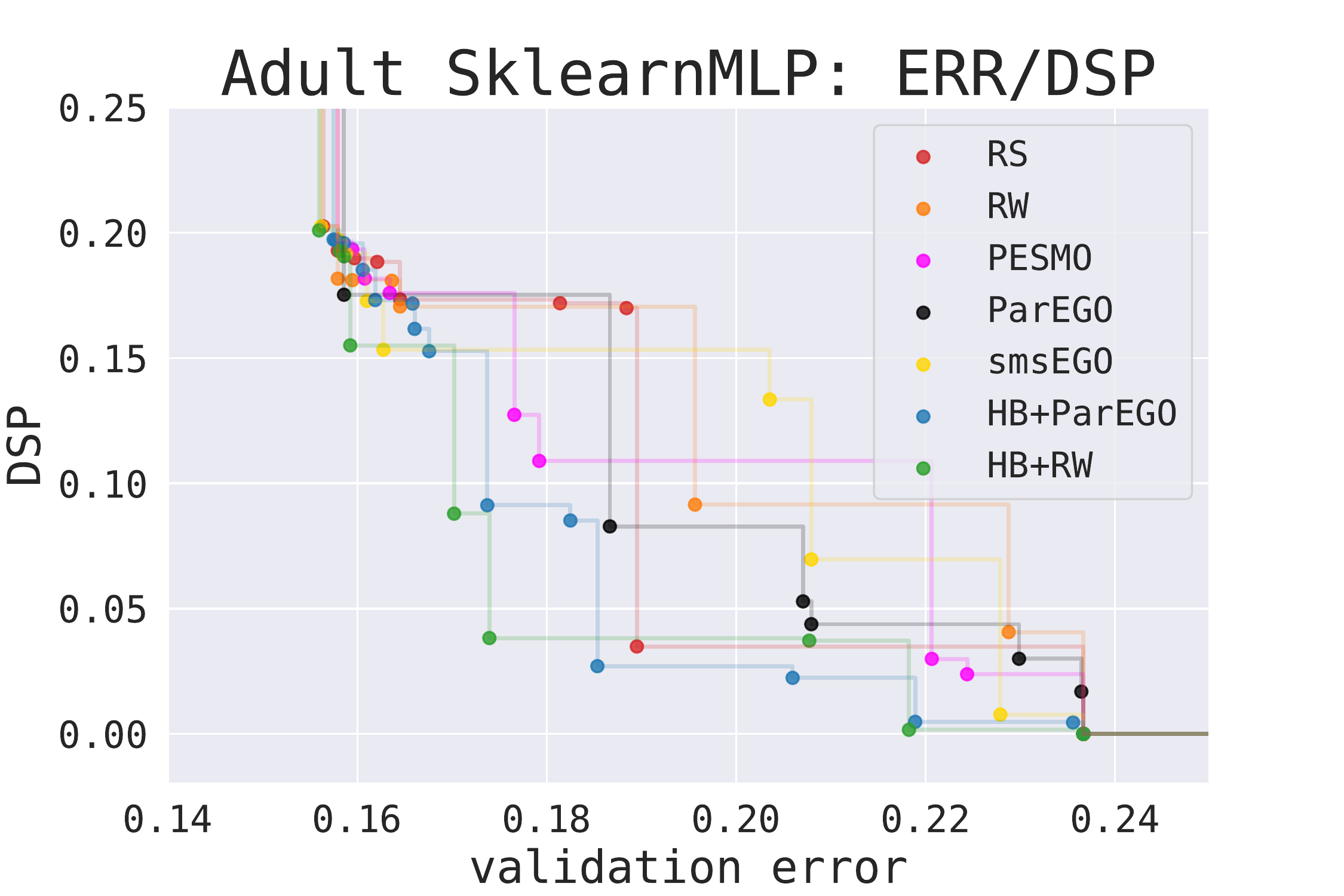}
        \includegraphics[width=0.49\textwidth]{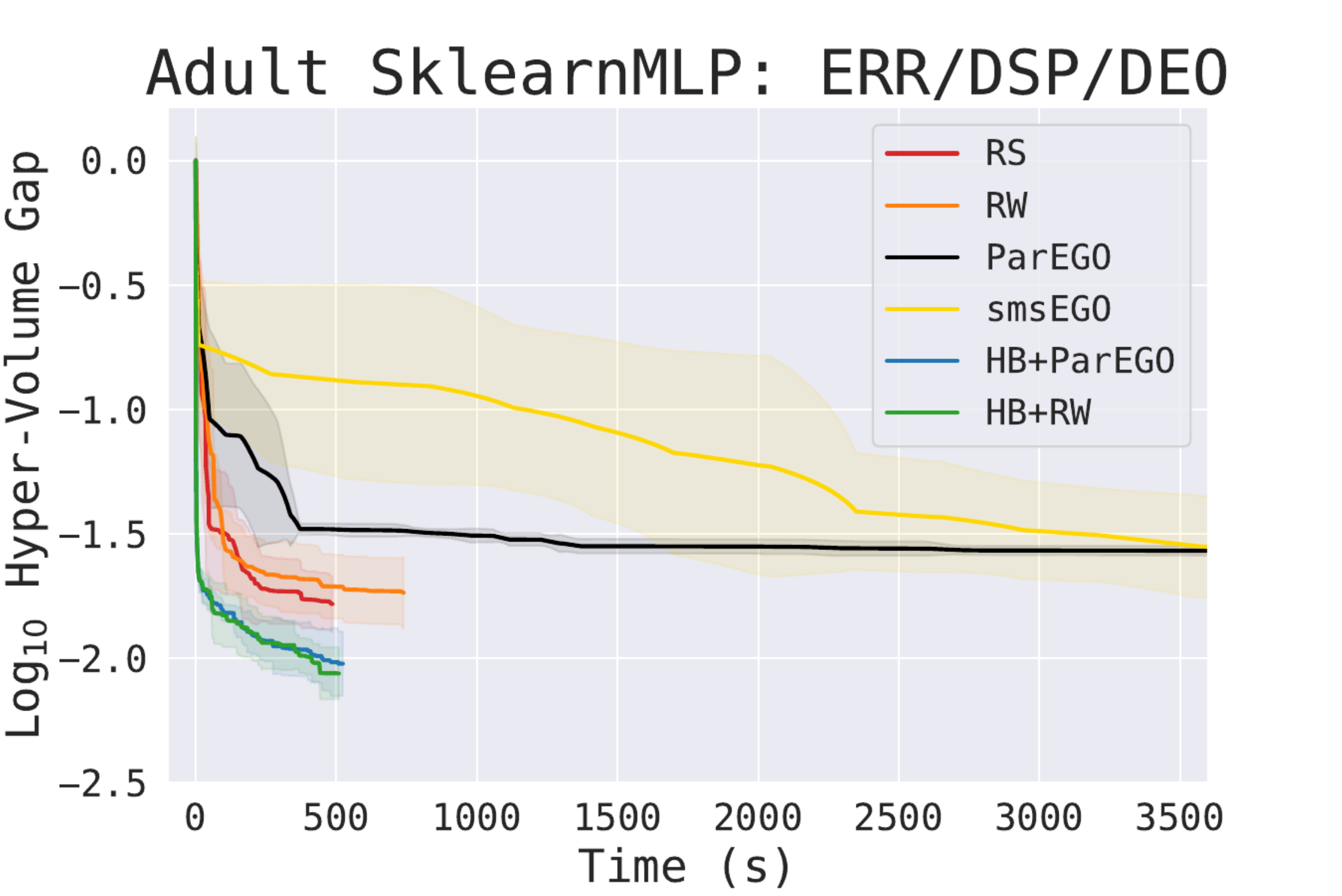}
        \includegraphics[width=0.49\textwidth]{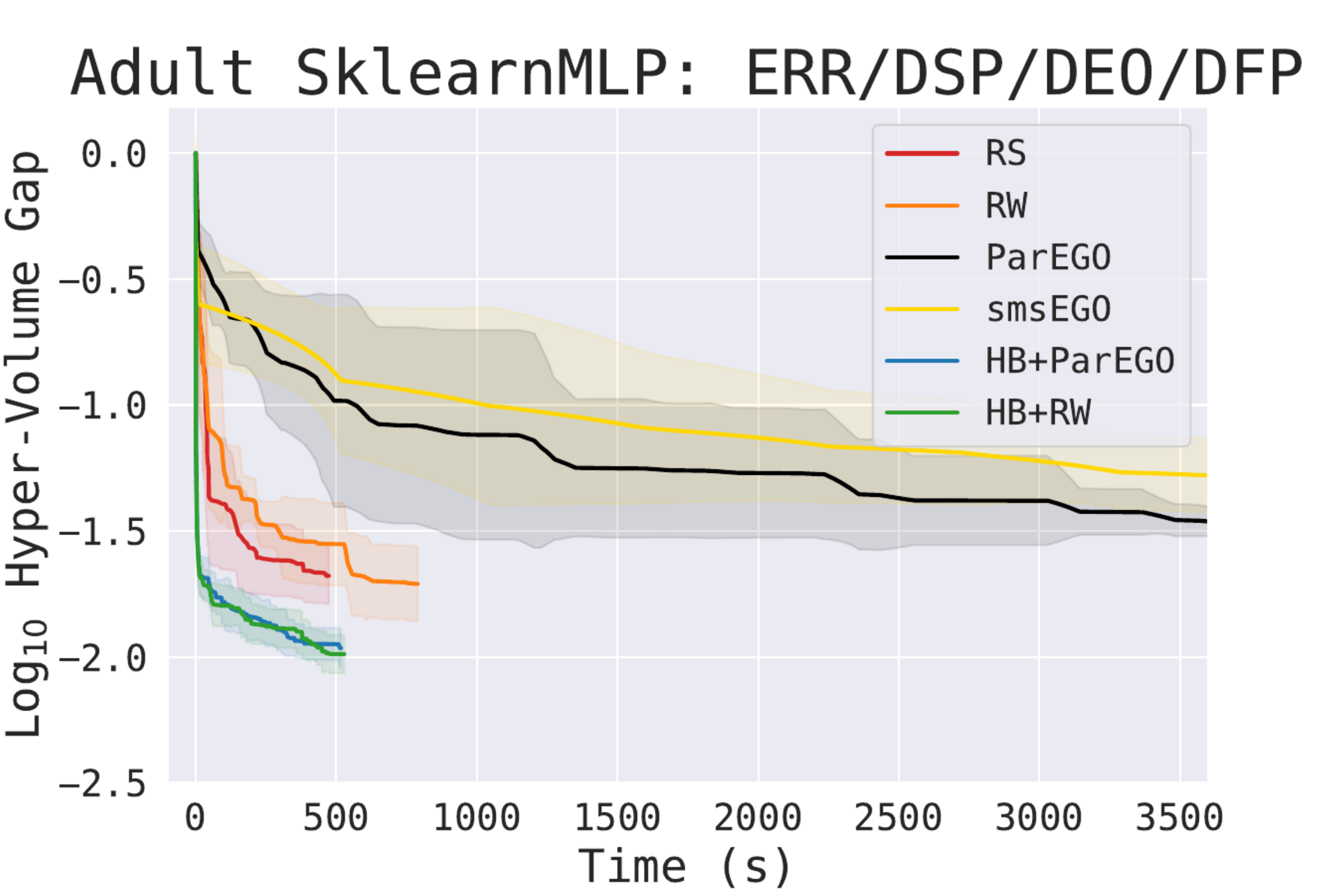}
    \caption{Difference in dominated hypervolume of the Pareto front  approximations returned by the different methods with respect to the combined front over time on Adult dataset. ERR, DSP, DEO and DFP objective are considered. The average and standard deviation for 5 random seeds is shown. For both scalarization schemes, the Hyperband method obtains Pareto front approximations with larger hypervolume in a shorter wall-clock time compared to multi-objective Bayesian optimization techniques. Random search is surprisingly competitive.}
    \label{fig:workshop-adult-mlp}
\end{figure}

A recent work~\cite{Schmucker2020:Multi} evaluated various MBO techniques as well as random search on multiple HPO tasks and artificial functions and investigated the use of multi-fidelity optimization for multi-objective HPO. With Figure~\ref{fig:workshop-adult-mlp} we include an excerpt of that study showing dominated hypervolume over time averaged over 5 random seeds for Sklearn MLP optimization on Adult datasets for 2 (ERR, DSP), 3 (ERR, DSP, DEO) and 4 (ERR, DSP, DEO, DFP) objective fairness settings. While~\cite{Schmucker2020:Multi} allowed the Hyperband methods to use multiple parallel workers we restrict each method to serial execution. One can observe that on these tasks random search can yield comparable sample complexity while causing minimal computational overhead. Unlike GP-based iterative methods, random search is easy to parallelize. Both are desirable properties for HPO algorithms intended to be run on large-scale systems. This motivates the use of multi-fidelity techniques such as Hyperband~\cite{Li2017Hyperband} and ASHA~\cite{Li2018:Massively} as natural extensions of random search for designing multi-objective HPO algorithms.

